\documentclass[10pt,twoside,journal]{IEEEtran}      % Use this line for a4
\IEEEoverridecommandlockouts                              % This command is only
\usepackage{graphics} % for pdf, bitmapped graphics files
\usepackage{mathptmx} % assumes new font selection scheme installed
\usepackage{mathrsfs}

\usepackage{times} % assumes new font selection scheme installed
\usepackage{amsmath} % assumes amsmath package installed
\mathchardef\mhyphen="2D
\usepackage{amssymb}  % assumes amsmath package installed
\usepackage{subfloat}
\usepackage[dvips]{graphicx}
\usepackage{cases}
\usepackage{floatrow}
\usepackage{etoolbox}
\makeatletter
\patchcmd{\@makecaption}
  {\scshape}
  {}
  {}
  {}
\makeatletter
\patchcmd{\@makecaption}
  {\\}
  {.\ }
  {}
  {}
\makeatother
\def\tablename{Table}

\usepackage[ruled,noend]{algorithm2e}
\usepackage[caption=false]{subfig}
\usepackage{multirow}
\usepackage{booktabs}
\usepackage{siunitx}
\usepackage{blindtext}
\usepackage{cite}
\usepackage{hyperref}
\hypersetup{hidelinks,
	colorlinks=true,
	allcolors=black,
	pdfstartview=Fit,
	breaklinks=true}

\IEEEoverridecommandlockouts                              % This command is only needed if 

\title{\LARGE \bf
Relay Hindsight Experience Replay: Self-Guided Continual Reinforcement Learning for Sequential Object Manipulation Tasks with Sparse Rewards}

\author{
Yongle~Luo$^{1,2}$,
Yuxin~Wang$^{1,2}$,
Kun~Dong$^{1,2}$,
Qiang~Zhang$^{1}$,
Erkang~cheng$^{1}$, \IEEEmembership{Member,~IEEE}, 
Zhiyong~Sun$^{1}$, \IEEEmembership{Member,~IEEE}, 
and Bo~Song$^{\#1}$, \IEEEmembership{Member,~IEEE}
\thanks{This work is supported in part by the grant of KRDP of Anhui Province (201904a05020086), NSFC(61804100, 61973294, 61806181) and CAS (GJTD-2018-15).}
\thanks{$^1$Institute of Intelligent Machines, Hefei Institutes of Physical Science, Chinese Academy of Sciences, Hefei, 230031, China}
\thanks{$^2$University of Science and Technology of China, Hefei 230026, China}

\thanks{$^{\#}$Corresponding author, email: {\tt\small songbo@iim.ac.cn}}

\thanks{The project is available at \href{https://github.com/kaixindelele/RHER/}{\emph{http://github.com/kaixindelele/RHER}}}.
}

\begin{document}
\bibliographystyle{IEEEtran}

\maketitle
\thispagestyle{empty}
\pagestyle{empty}

%%%%%%%%%%%%%%%%%%%%%%%%%%%%%%%%%%%%%%%%%%%%%%%%%%%%%%%%%%%%%%%%%%%%%%%%%%%%%%%%
\begin{abstract}
% 强化在稀疏奖励下的探索仍然是一个具有挑战性的问题。尤其是对于复杂的序列物体操作任务来说，在没有完成所有的子任务之前，智能体都只能获得负的奖励，这将导致极高的探索复杂度。为了解决这个问题，我们提出RHER算法。由于子任务目标的简化，利用HER可以帮助智能体快速学会最简单的任务。
Exploration with sparse rewards remains a challenging research problem in reinforcement learning (RL). Especially for sequential object manipulation tasks, the RL agent always receives negative rewards until completing all sub-tasks, which results in low exploration efficiency. To solve these tasks efficiently, we propose a novel self-guided continual RL framework, Relay-HER (RHER). RHER first decomposes a sequential task into new sub-tasks with increasing complexity and ensures that the simplest sub-task can be learned quickly by utilizing Hindsight Experience Replay (HER). Secondly, we design a multi-goal \& multi-task network to learn these sub-tasks simultaneously. Finally, we propose a Self-Guided Exploration Strategy (SGES). With SGES, the learned sub-task policy will guide the agent to the states that are helpful to learn more complex sub-task with HER. By this self-guided exploration and relay policy learning, RHER can solve these sequential tasks efficiently stage by stage. The experimental results show that RHER significantly outperforms vanilla-HER in sample-efficiency on five single-object and five complex multi-object manipulation tasks (e.g., Push, Insert, ObstaclePush, Stack, TStack, etc.). The proposed RHER has also been applied to learn a contact-rich push task on a physical robot from scratch, and the success rate reached 10/10 with only 250 episodes. 
\end{abstract}

\begin{IEEEkeywords}
Deep reinforcement learning, Robotic manipulation, Continual Learning, Residual Policy Learning
\end{IEEEkeywords}

%%%%%%%%%%%%%%%%%%%%%%%%%%%%%%%%%%%%%%%%%%%%%%%%%%%%%%%%%%%%%%%%%%%%%%%%%%%%%%%%
\section{Introduction}
% 参考acder的描述：强化在很多序列决策任务中都取得了很大的突破。领域从打游戏，到下围棋，尤其是大量的机器人操作任务，比如抓取，开门，和序列物体操作任务。
Deep reinforcement learning (RL) has made many breakthroughs in various sequential decision-making problems, ranging from playing Atari \cite{atari}, Go \cite{go} to control tasks \cite{ppo}, especially for a number of robotic manipulation tasks such as grasping \cite{MT-opt}, door opening \cite{door} and object manipulation \cite{HER1}. 

However, it is still challenging to learn policies for sequential object manipulation tasks with sparse rewards. Due to receiving negative reward signals (e.g., r~=~-1) until completing tasks, the agent can not distinguish which action is better according to these identical negative rewards. Few valuable samples contribute to guiding policy optimization \cite{sparse-gradient}, which results in exponential sample complexity \cite{SHER}.

% HER通过把失败的目标，修改为回合中已经完成的目标，产生非负奖励缓解稀疏奖励。对于简单的reach任务，智能体确实可以通过这种方法得到效率的提升。
With HER \cite{HER1}, the agent can generate non-negative rewards (e.g., r~=~0) by goal relabeling strategy to alleviate the negative sparse reward problem, even if the agent did not complete the task. For simple reach tasks, the agent can benefit from HER in efficiency.

%To alleviate the negative sparse reward problem, the Hindsight Experience Replay (HER) \cite{HER1} method generate non-negative rewards by relabeling the failure desired goal with the goal achieved in the episode. For simple reach tasks, the agent can benefit from HER in efficiency.

\begin{figure}[h]
	\hspace*{-2mm}
	\includegraphics[width=0.99\textwidth]{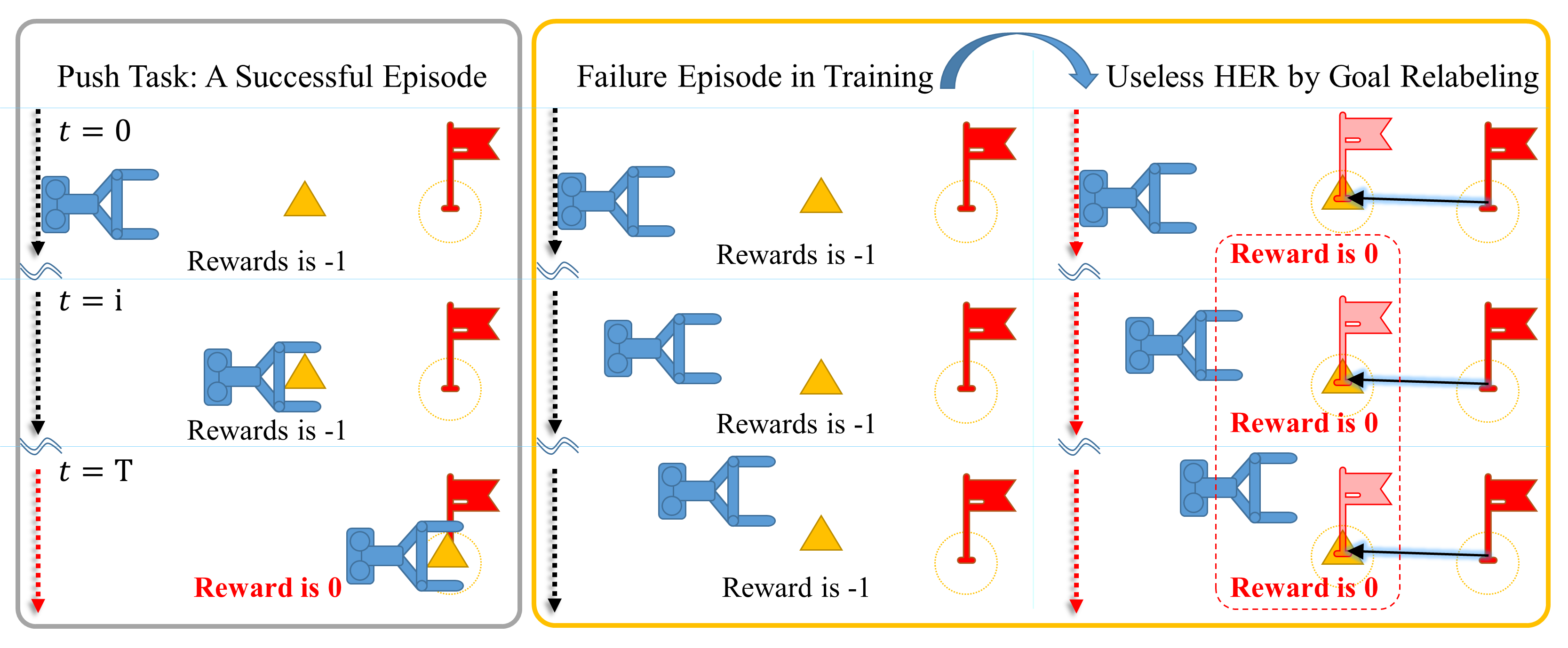}
	\caption{Illustration of the problem of non-negative sparse rewards with HER. For a typical sequential task, push task, the agent fails to push the object to the desired position, and even fails to change the object position. So all original rewards are -1, and all hindsight rewards are 0, the latter can also be regarded as a kind of sparse reward problem, but with non-negative rewards.}
	\label{Fig-HER_push}
\end{figure}

%然而，对于复杂的序列物体操作任务，智能体需要先完成一些子任务，才能实现最终的目标结果。当智能体无法完成子任务的时候，利用HER算法会引入另外一种稀疏奖励问题。
%即当智能体无法改变ag的时候，HER的产生的样本和原始的样本都无法让智能体分辨，哪个动作更好。我们把这种现象称之为nnsr问题。
%总的来说，当智能体的能力还不够改变最终目标的结果的时候，使用HER这种好高骛远的行为，并不会给策略带来提升，甚至会降低网络探索能力。
%把影响不了结果的现象，放到句首。
But for complex sequential object manipulation tasks, the agent still suffers from low sample efficiency with HER, due to another implicit sparse reward problem. For these sequential manipulation tasks, the agent needs to accomplish each sub-task before it can achieve the final desired goal. As shown in Fig. \ref{Fig-HER_push}, in a toy push task, the agent needs to reach the object and push it to a desired position. But if the agent is unable to change the object position, the achieved goals (i.e., the object position) are identical in the whole episode. Thus, the hindsight goals are also identical, which means that all hindsight rewards are non-negative. The agent can not distinguish which action is better from these original samples or hindsight samples. We call this phenomenon as non-negative sparse reward (NNSR) problem. In general, when the agent's ability is not enough to change the outcome of complex tasks, it still uses HER to change the goals to fantasize about success. This kind of ambitious behavior will not only be unhelpful for policy improvement, but even reduce the exploration ability of network \cite{reward-shift} due to a large number of valueless non-negative hindsight samples.

%Illustration of the problem of low sample efficiency with HER for manipulation tasks.

% 为了解决这个NNSR的问题，智能体必须要能够影响对应的ag.所以我们提出了自引导的RHER算法。首先它通过目标空间编码技术实现不同难度的目标同时学习，先从简单的、能直接改变ag的任务开始学，利用HER产生有价值的非负奖励.其次，我们提出随机混合探索策略实现自我引导探索，利用已经学会的简单任务引导智能体，到达能够高概率改变更复杂任务的状态。

To solve these complex sequential tasks with sparse rewards more effectively, the agent needs to affect the corresponding achieved goals, which will alleviate the above implicit NNSR problem by HER. So we propose a novel self-guided continual RL framework, Relay-HER (RHER). Firstly, RHER decomposes the whole sequential task into multiple stages (or steps) and then combines them to new sub-tasks with increasing complexity. Secondly, inspired by one-hot encoding, a multi-goal \& multi-task network can be designed by a goal space encoding technique. With this structure, the agent can learn multiple sub-tasks with different complexity goals simultaneously. Third, the agent generates valuable non-negative samples for the accessible stages by using HER. Finally, a Self-Guided Exploration Strategy (SGES) is proposed. With SGES, the agent will be guided by the learned simpler policy to states that can change the achieved goal of a more complex sub-task with high probability. In other words, the agent can extend the accessible stages quickly by SGES. More efficient exploration is realized via self-guidance without human demonstrations or pre-defined controllers. The complex sequential tasks can be solved efficiently by this relay policy learning stage by stage.

The contributions of this study can be summarized as follows: 
% 针对问题，提出方法，实现了效果；
% 针对问题，提出方法，实现了效果；
% 实验结果表明；
% 针对稀疏奖励下的复杂序列操作问题，我们提出了自我引导的持续强化学习算法，RHER，可以一步一步解决。

(1) For common complex sequential manipulation tasks with sparse rewards, this paper develops an elegant and sample-efficient self-guided continual RL framework, RHER.

(2) To achieve self-guided exploration, we propose a multi-goal \& multi-task network to learn multiple sub-tasks with different complexity simultaneously.

(3) The proposed RHER method is more sample-efficient than vanilla-HER and state-of-the-art methods, which are validated in the standard manipulation tasks from the OpenAI Gym \cite{gym}; 

(4) To verify that the RHER is suitable for common sequential object manipulation tasks, we conduct three extra typical single-object tasks, five more complex multi-object tasks, and even a physical robot task.

\label{Section I}

\section{Related Work}
Many studies try to deal with complex sequential object manipulation tasks under sparse rewards for more efficiency.

\subsection{HER-based methods}
HER paves a promising path toward the sparse reward problem. Through the Universal Value Function Approximators (UVFA) \cite{UVFA} with neural networks, it can even be generalized to the unseen actual goal. HER has extremely high efficiency in handling long-horizon reaching tasks \cite{HER1, HER2}, but it is not efficient in sequential manipulation tasks. 

In order to improve the sample efficiency of HER, many variants of HER have been proposed: \textbf{1)} curriculum learning to maximize the diversity of achieved goals \cite{CHER, SHER}; \textbf{2)} providing demonstrations \cite{HER-Demo}, \textbf{3)} generating more valuable desired goals \cite{GHER2}; \textbf{4)} curiosity-driven exploration \cite{li2020acder}; \textbf{5)} model-based RL methods \cite{IHER}. These methods have improved the performance of HER to a certain degree, but none of these methods directly solve the NNSR problem.

The problem of identical achieved goals of HER is firstly reported in \cite{FHER}, the solution of which is to directly filter those `biased' transitions with ${r(s_{t-1}, a_{t-1}, g^{'})=0}$. This method improves the sample efficiency compared to the vanilla-HER for three throwing tasks. However, it is not suitable for the common manipulation tasks when the goal space is within the workspace because these transitions can “teach” the agent not to destroy the achieved goal. The problem is further discussed in \cite{SHER}. The researchers propose a SHER algorithm that learns a reaching policy first and then transfers the reaching policy to a more complex sub-task. However, SHER can not learn multiple sub-tasks simultaneously, so it has no self-guided exploration.

\subsection{Hierarchical RL with HER}
To simplify the sequential manipulation tasks, many works combine hierarchical RL (HRL) with HER to solve these sub-tasks \cite{HAC, HRL}. Specifically, \cite{HRL} effectively solves the stacking problem of three blocks in different orders by abstract demonstrations, $block$-$gripper$-$informed$~$goals$ and auto-adjusting exploration strategy. However, these HRL methods do not explicitly use the learned policy to guide a more complex policy during exploration, so the efficiency has not been improved enough. In addition, the $block$-$gripper$-$informed$~$goals$ is not suitable for some tasks, in which the target position of the gripper can not be set in advance, such as pushing tasks.
%结果：效率没有得到充分的提升，但是没有做对比实验，这样描述可能不太好。

% 改进该怎么描述？
\subsection{Continual learning and Continual-RL}
Different from the aforementioned HER-based or HRL-based algorithms, this study adopts the continual RL scheme to solve sequential robotic tasks. CL has drawn increasing attention from the RL community over the last few years. 

Current mainstream CL has three solutions to keep the previous tasks from forgetting \cite{CL-HyperNet}. The first is the regularization-based scheme \cite{CL-regular1, CL-regular2}. The second is the modular approaches \cite{CL-mod1, CL-mod2}, for example, by designing a task-conditioned architecture, it can accommodate both previous and new tasks \cite{CL-HyperNet}. The third is the memory-based method which needs to revisit previous task data when training for a new task \cite{CL-memory1, CL-memory2}. The proposed RHER use a novel task-condition architecture and an inherent experience buffer of RL to avoid forgetting.

For object manipulation tasks, few methods use Deep RL with CL. \cite{MB-CL} trains a hyper-network to get dynamics models with different dynamics tasks. To sidestep the exploration dilemma, this work places the object close to the gripper. \cite{Lifelong} filters out samples not suitable for the target task, then relabel the remaining samples to pre-train a separate policy for a new task. In addition, both articles use a heuristic reward function.

In contrast to the above continual RL methods, RHER can share data across all sub-tasks by goal relabeling, nor do we need to design a heuristic reward function for each task, which means that it can be adapted to various tasks and scenarios with little effort. 

\label{Section II}

%BACKGROUND
\section{Preliminaries}
% the reason why HER cannot solve sparse reward problems for two-stage tasks efficiently will be briefly analyzed.
This section introduces the background of RHER, including goal-conditioned RL and HER.

\subsection{Goal-conditioned RL}
According to the HER methods, manipulation tasks can be modeled as a finite-horizon, discounted Markov Decision Process (MDP). For a goal-conditional RL, the whole state $S$ contains not only the observation $S_o$, but also a desired goal $S_{dg}$, where $S_{dg}{\in}S_o$. The policy ${\pi}_{\theta}(a_t|s_{o_t},s_{dg_t})$ is usually a neural network model that maps the $t$ step state $s_t=(s_{o_t},s_{dg_t})$ to the action $a_t{\in}A$. The MDP has some other necessary components: a discount factor $\gamma{\in}(0, 1)$, horizon length $T$ and a reward function $r{\colon}S{\times}A{\to}R$. The goal of the agent is to acquire an optimal policy ${\pi}_{\theta}(a|s_o,s_{dg})$that maximizes the expected sum of discount rewards.

It is noted that any off-policy actor-critic algorithms or variants of HER can be used in the proposed RHER. As for the classic DDPG \cite{DDPG} algorithm, there are two main parts, an actor neural network denoted as ${\pi}_{\theta}(a|s_o,s_{dg})$, for generating actions, and a critic network denoted as $Q_{\phi}(s_o,s_{dg},a)$, for evaluating the action performance. To stabilize the learning process, there is also a set of target actor ${\pi}_{\theta}^{'}$ and target critic $Q_{\phi}^{'}$ with the same structure but delayed update. The critic network parameter $\phi$ is learned by minimizing the Bellman error, and the loss function is defined as (\ref{eq-closs}),
\begin{equation}\label{eq-closs}
{Loss}_{\phi}^{(j)}=|Q_{\phi}(s_{o_t},s_{dg_t}^{(j)},a_t)-y_t|^2,
\end{equation}
where the target $y_t=r_t+{\gamma}Q_{\phi}^{'}(s_{o_{t+1}},s_{dg_{t+1}},a_{t+1})$; $a_{t+1}$ is generated by ${\pi}_{\theta}^{'}(s_{o_{t+1}},s_{dg_{t+1}})$. To extend multi-task setting, let $s_{dg_t}^{(j)}$ denote the desired goal of $j^{th}$ task.

Based on the Q function, the policy network parameter
$\theta$ is trained using the gradient descent method with the loss represented as (\ref{eq-aloss}),
\begin{equation}\label{eq-aloss}
{Loss}_{\theta}^{(j)}=-{\mathbb{E}}_{s_{o_t},s_{dg_t}^{(j)}} Q[s_{o_t},s_{dg_t}^{(j)},{\pi}_{\theta}(s_{o_t},s_{dg_t}^{(j)})].
\end{equation}

Usually, the reward feedback is sparse, represented as (\ref{eq-rew}),
\begin{equation}\label{eq-rew}
{r(s_{o_{t+1}},s_{dg_t},a_t)}=-\mathbb{I}(s_{ag_{t+1}}=s_{dg_t}),
\end{equation}
where $\mathbb{I}()$ is the indicator function, and $s_{ag_{t+1}}{\in}S_{o_{t+1}}$ is the achieved goal in the next state. This reward function indicates whether the current task was completed.

\subsection{HER Goal Relabeling}
To use HER, we specify the following concepts:

\begin{itemize}
	\item $\mathbf{desired~goal}$: the goal of the current episode, specifically, which is usually a target position.\par
	\item $\mathbf{achieved~goal}$: the goal achieved in the current state. It represents the end-effector position in a reaching task, whereas the object position in a manipulation task. 
	\item $\mathbf {hindsight~goal}$: in experience replay, the HER method uses the future achieved goal in the episode as a hindsight goal.\par

\end{itemize}

According to vanilla-HER, with a probability of 0.8, we relabel the original desired goal $s_{dg_t}$ with another observation from a future time step of the same episode as (\ref{eq-her}):

\begin{equation}\label{eq-her}
	{s_{dg^h_t}}=s_{ag_{t+k}}, 1\leqslant k\leqslant T,
\end{equation}

where the $s_{dg^h_t}$ is the hindsight goal for $t$ time step, $s_{ag_{t+k}}$ is the achieved goal for $t+k$ time step, and $T$ is the last time step of the episode.

%Training of Representation Model
\section{Methods}

% 本章节展示一个全新的，高效的，自我引导的，持续强化学习框架，RHER。对于序列操作任务来说，它通过四个模块的处理，实现了自我引导探索，可以快速学会完整的任务。
%它由四个部分组成的，按照传统的强化交互信息流可以参考图2.

\begin{figure}[h]
	\hspace*{-2mm}
	\includegraphics[width=0.99\textwidth]{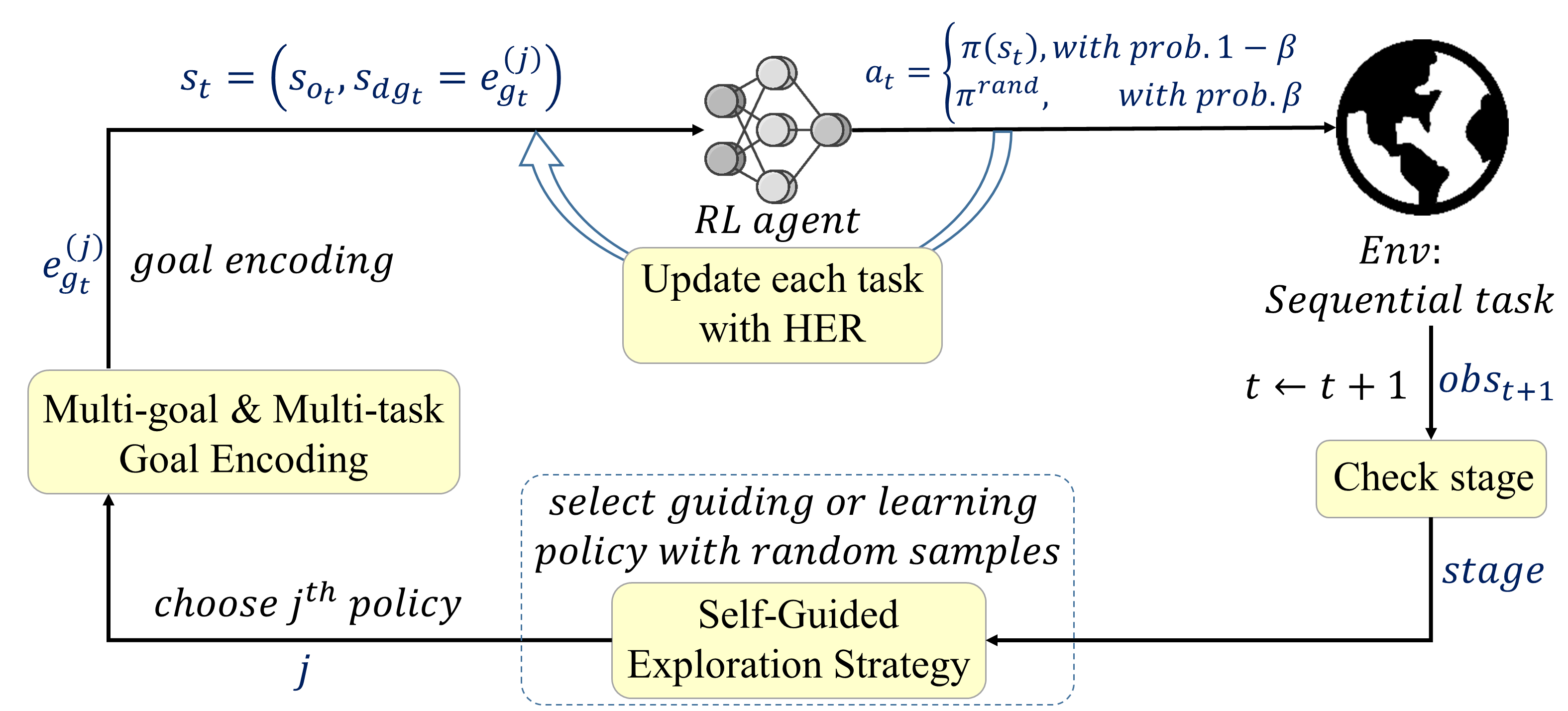}
	\caption{A diagram of RHER, of which the key components are shown in the yellow rectangles. This framework achieves self-guided exploration for a sequential task.}
	\label{Fig-RHER-overall}
\end{figure}

%详细介绍RHER方法，为了实现对序列操作任务的快速学习，我们首先
%这个方法实现了接力式学习和自我引导，
This section presents a novel self-guided continual RL framework, RHER. Four key components of RHER cooperate with each other to realize self-guided exploration, which enables the agent to quickly solve the complex sequential object manipulation tasks, as shown in Fig. \ref{Fig-RHER-overall}. 

%根据最初的设计逻辑，四个组件分别是：
%1）这个组件
According to the logical order of design, the four components are: \textbf{1)} The first component is to decompose the original sequential task into $M$ stages (or steps) and then combine them to $M$ sub-tasks with increasing complexity; \textbf{2)} The second component is to design a multi-goal \& multi-task RL model to deal with the above sub-tasks simultaneously; \textbf{3)} By utilizing HER, this component overcomes the problem of negative sparse rewards and shares data collected from different sub-tasks; \textbf{4)} To collect more valuable samples and alleviate the NNSR by HER for sequential tasks, a Self-Guided Exploration Strategy (SGES) is proposed. The SGES uses a learned policy to guide the agent quickly to more complex stages. Through this self-guided exploration, agents can learn sequential tasks stage by stage, similar to a relay.

% clever是不是要加得另说；
% 对于持续强化学习来说，四个的组合实现了稳定的知识前向传递，确保了简单策略也不会遗忘。
As a continual RL framework, the combination of these components realizes a stable knowledge transfer from a simple policy to a complex one and ensures that the simple policy will not be forgotten.

The full RHER algorithm is described in Algorithm\ref{algorithm1} and the details of these four parts are introduced as follows.

\begin{algorithm}
  \SetKwInOut{Input}{input}
Initialize the agent and replay buffer R\\
\For{$each~episode$}{
     Sample an observation dictionary\\
     \For{$each~time~step$}{
			 Check stage as (\ref{eq-stage}) \\
			 Sample an action with \textbf{SGES} as (\ref{eq-SGES})\\
      Execute the action and get new observation\\
      Store the transition\\           
     }
     \For{$each~gradient~step$}{
         \For{$each~sub$-$task$}{
              Sample a mini-batch episodes $B$ from R\\
              Replace the goal $e_g^{(j)}$ as (\ref{eq-eg-next})\\
              Perform optimization of (\ref{eq-closs}), (\ref{eq-aloss}) based on $B$ \\
         }
     }
}
  \caption{Relay Hindsight Experience Replay}\label{algorithm1}
\end{algorithm}

%% 相比原始的HER方法，完整的方法只引入了两个超参数。
%The whole method introduces two main insensitive hyper-parameters in contrast to the vanilla-HER methods, one is in the sub-task segmentation, the threshold of the distance object needs to be set in advance, and the other is the guidance ratio in SGES. 

\subsection{Task Decomposition and Rearrangement}
% 为了搜集到更高质量的样本，智能体必须要能够改变对应的ag. 而对于普通的物体操作任务来说，原始任务的ag一般很难在一开始就被智能体改变。因此我们根据物体操作任务的特性，将N个物块的任务，分解成2N个阶段。如图3的task description所说，第2i-1个阶段是到达第i个物块，且要保证前i-1个物体要在指定位置，第2i个阶段是要把第i个物体move到指定位置。

% 对于常见的序列物体操作任务来说，最终的目标，在开始的时候，很难被智能体改变。为了搜集到有价值的样本，智能体必须要搜集到被它改变结果的样本。因此我们需要将序列任务分成M个阶段，重组成难度依次递增的任务，让智能体能够从最简单的任务开始学起。
\begin{figure}[h]
	\hspace*{-2mm}
	\includegraphics[width=0.9\textwidth]{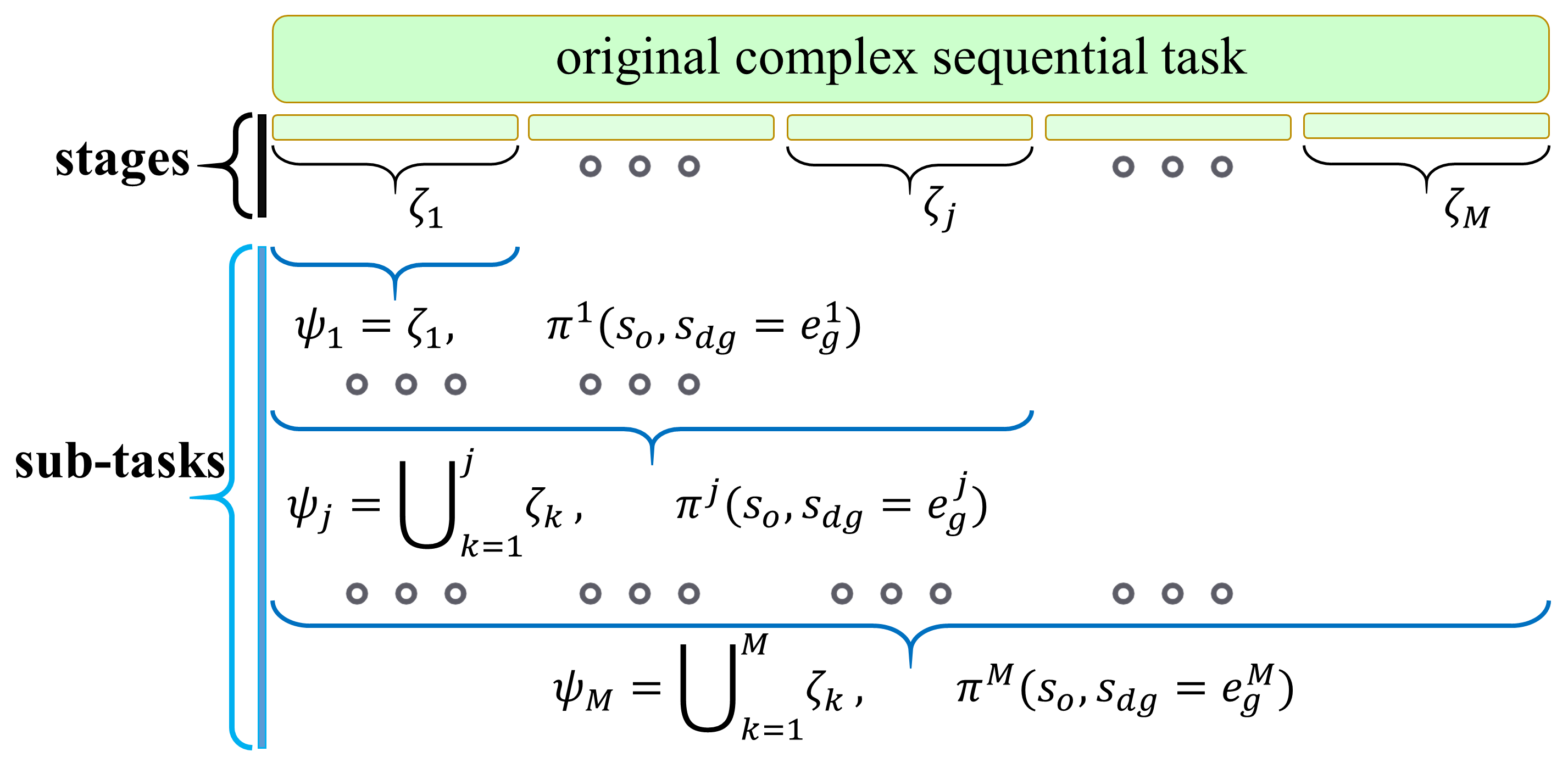}
	\caption{Sequential task decomposition and rearrangement.}
	\label{Fig-RHER-task}
\end{figure}

In this paper, we decompose the original sequential task into $M$ crucial stages (\{$\zeta_j|j=1,2...M$\}) and rearrange them to $M$ new sub-tasks (\{$\psi_j|j=1,2...M$\}) with increasing complexity. As shown in Fig. \ref{Fig-RHER-task}, $\psi_j$ is given by

\begin{equation}\label{eq-task}
	{\psi_j}={\cup}_{k=1}^{j}\zeta_k,~~~j\in{\{1, 2, ... , M\}},	
\end{equation}

The corresponding policy of $\psi_j$ is denoted as $\pi^j$. Note that if the current stage is $\zeta_j$, which means that the agent has solved the stages from $\zeta_1$ to $\zeta_{j-1}$. In addition, the policy $\pi^j$ can handle the stages from $\zeta_1$ to $\zeta_j$, and $\pi^1$ just can deal with the $\zeta_1$.

If there are $N$ objects to manipulate in order, this sequential task can be divided into $2N$ stages (i.e., $M=2N$), according to the natural order of task. To be specific, the aim of $(2i-1)^{th}$ stage $\zeta_{(2i-1)}$ is to reach the $i^{th}$ object and prepare for the $(2i)^{th}$ stage, and the $(2i)^{th}$ stage $\zeta_{(2i)}$ is to manipulate the $i^{th}$ object to reach its target goal. 

% 这里我们需要先定义一个超参数，即末端到物块的距离，如图SGES所示，具体的实验见第六章的C。
The distance from the end-effector to object $d$ needs to be defined in advance (ablation experiments are shown in Section \ref{Section VI.C}). The current stage index $j$ can be determined by (\ref{eq-stage}),
%%%%%%%%%%%%%%%%%%%%%%%%%%%%%%%%%%%%%%%%%%
\begin{equation}\label{eq-stage}
	j=\left\{
	\begin{aligned}
		& 2*i-1 &  \text{if~}\:dist_i\:\:>\:d\\
		& 2*i & \text{if~}\:dist_i\:\:\le \:d\\
	\end{aligned}
	\right
	.
\end{equation}

where $dist_i$ is the distance between the end-effector and the $i^{th}$ object. If $N > 1$, it means that the first $(i-1)^{th}$ objects have been solved.

According to (\ref{eq-task}), we rearrange the above stages into $M=2N$ new sub-tasks. Considering that some manipulations, such as pushing and sliding, require continuous decision-making, this task rearrangement is necessary. A detailed ablation study will be conducted to discuss the relationship between the sub-tasks and stages (see Section \ref{Section VI.B}). 
%The relationship can be formulated as (\ref{eq-task}),
%\begin{equation}\label{eq-task}
%	{\psi_j}={\cup}_{k=1}^{j}\zeta_k,~~~j\in{\{1, 2, ... , M=2N\}},	
%\end{equation}

%From (\ref{eq-task}), we can see that the difficulty of sub-tasks increases gradually with the increase of sub-task index $j$.
%图三的任务介绍是一个典型的例子，它描述了三个物体的任务切分。

For a more concrete example, in a two-block pushing task, the $2^{nd}$ stage $\zeta_2$ is that the end-effector has reached the $1^{st}$ object, but neither of the blocks has reached the specified location; the $3^{th}$ stage $\zeta_3$ represents that the $1^{st}$ object has been moved to its target, but the second one hasn't, and the end-effector is far from the second ones. The $3^{th}$ sub-task $\psi_3$ means that: 1) reach the $1^{st}$ block, 2) push the $1^{st}$ block to the desired location, 3) leave the $1^{st}$ block and keep it in position, then reach the position of $2^{nd}$ block. 
\label{Section IV.A}

% Without loss of generality, this paper focuses on the single object tasks (N=1). Specifically, the $\psi_1$ here means controlling the end-effector to reach the object, termed $\mathit{reaching{\mhyphen}task}$, as shown in Fig. \ref{Fig2}(b). The achieved goal here is the gripper's position, named $\mathit{grip{\mhyphen}pos}$; the desired goal is the coordinate of the object, denoted as $ag^1$. As for the $\psi_2$, it is essentially the original task, namely $\mathit{target{\mhyphen}task}$, where the achieved goal is the position of the object, represented as $\mathit{obj{\mhyphen}pos}$, and the desired goal is denoted as $dg^1$. Finally, $\pi^1$ and $\pi^2$ correspond to the reaching policy and the target policy, respectively (as shown in Fig. \ref{Fig2}(b)). Of course, this task setting can also be easily extended to the tasks of multiple objects.

%For many object manipulation tasks, the regular HER methods confront the problem of non-negative sparse rewards. To tackle these complex long-horizon tasks, a common idea is to decompose them. 

%After the decomposition is done, the next step is to form new tasks. There are two strategies, one is that one task corresponds to one stage, and the other is that each task processes one more stage based on the previous task. Considering the dynamic continuous tasks such as pushing and sliding, the latter is more stable, and for the pick-related tasks, the first strategy is easier for learning. Detailed ablation experiments will be conducted in Section \ref{Section VI.B}.\label{Section IV.A}

\subsection{Multi-goal \& Multi-task RL Model}

When the original complex task is decomposed into 2N new tasks with increasing complexity, an intuitive solution is to use the idea of CL to resolve these sub-tasks in order. The objective of CL is to learn new knowledge of the current task continuously, while the performance on previous tasks is not forgotten. In this manner, a network structure that can deal with multiple sub-tasks is required. The standard multi-task RL policy can be denoted as (\ref{eq-mgmt-pi}):

\begin{equation}\label{eq-mgmt-pi}	 {\pi}_{\theta}(a|s_o,s_{dg}^{(j)}=e_g^{(j)}),
\end{equation}

where $a{\in}A$ represents the action comprised of a 4-D vector for controlling 3-D position and 1-D opening of the gripper in this study; $s_o{\in}S$ represents the system states comprised of Cartesian positions, the linear velocity of the gripper and the position and velocity information of the object; $s_{dg}^{(j)}$ represents the desired goal of $\psi_j$; $e_g^{(j)}$ represents the specific goal space encoding of $s_{dg}^{(j)}$.

\begin{figure}[h]
	\hspace*{-2mm}
	\includegraphics[width=0.95\textwidth]{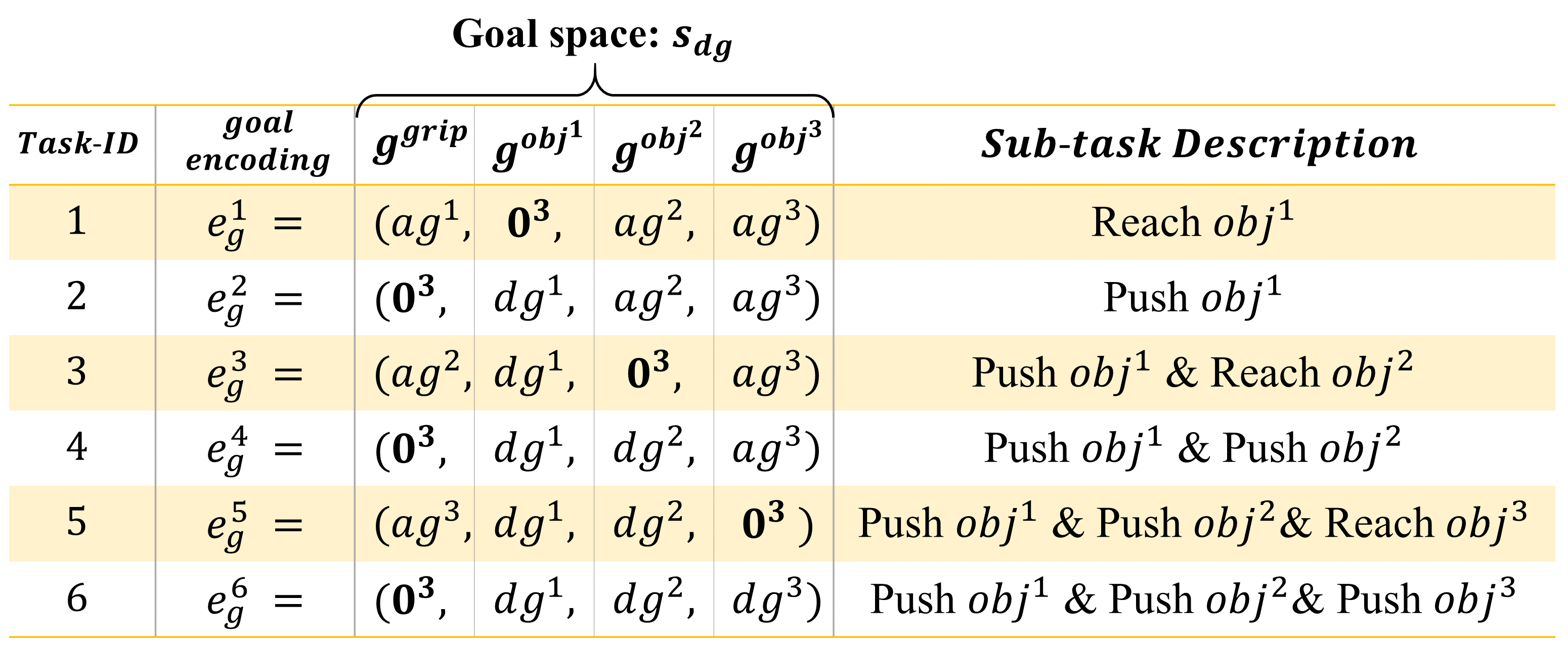}
	\caption{An example of multi-goal \& multi-task goal space encoding for a three-object push task. $ag^i$ is the achieved goal of $obj^i$ (i.e., the current position of $obj^i$); $dg^i$ is the desired goal of $obj^i$; $\textbf{0}^3$ is 3-D zero vector which plays a key role in the task identity.}
	\label{Fig-RHER_goal_encoding}
\end{figure}

% 很明显，设计一个优雅且有效的目标编码是后续自我引导探索的关键前提。然而普通的多任务是靠one-hot编码来作为输入实现的，这意味着编码任务是离散的，并不适合目标空间连续的GCRL。不过仍然受到one-hot encoding的启发，我们use one-hot to zero-padding，用做区分任务种类。
Designing an elegant and effective $e_g^{(j)}$ is a key prerequisite for the following self-guided exploration. The traditional multi-task network depends on a one-hot task ID with the state as inputs \cite{MT-opt, CL-mod1}. However, one-hot goal representations are not suitable for continuous goal space. 

To design a multi-goal \& multi-task network, inspired by the one-hot encoding, we use the zero-padding encoding (i.e., a 3-D zero vector $\textbf{0}^3$.) to identify tasks. 

% 对于N个物体的操作任务,如果只考虑前i个物块到指定的位置，且不动后面N-i个物体，那么\psi_j的编码e_g^j的通式表述为：
%e_g^j = [ag^i||{dg}^{1}...{dg}^{i-1}||\textbf{0}^3||ag^{i+1}...ag^N],   && \text{if~}j=2i-1,
%e_g^j = [\textbf{0}^3||{dg}^{1}...{dg}^{i-1}||{dg}^i||ag^{i+1}...ag^N], && \text{if~}j=2i,

% 先从本文尝试解决的最复杂的三物体任务为例子，如图3所示，通过[5][6]一共拆解出了6个任务，其中有三个包含了reach到指定位置，
We take a complex three-object task as an example, as shown in Fig. \ref{Fig-RHER_goal_encoding}. In this sequential task, three objects are denoted by $obj^i$ ($i = 1,2,3$), the corresponding position of these objects are denoted by $ag^i$ ($i = 1,2,3$), and the corresponding goal are denoted by $dg^i$ ($i = 1,2,3$). To perform the task, six sub-tasks, i.e. \{$\psi_j|j=1,2...6$\} are defined through (\ref{eq-task}). The goal space $s_{dg}^j$ consists of $g^{grip}$, $g^{obj^1}$, $g^{obj^2}$ and $g^{obj^3}$, which are the goals of gripper and three object respectively. The aim of $\psi_1$ is to reach $obj^1$, so the goal of the gripper, i.e. $g^{grip}$ is the current position of the $obj^1$, i.e. $ag^1$. Noted that, because the agent can not affect the $obj^1$ in this stage, the goal of $obj^1$, $g^{obj^1}$, is replaced by $\textbf{0}^3$, which also can distinguish the present task from others. In addition, $g^{obj^2}$ and $g^{obj^3}$ are $ag^2$ and $ag^3$, respectively, which mean that $\psi_1$ should not change the current position of these objects. Another typical sub-task is $\psi_4$, the aim of $\psi_4$ is to push $obj^1$ and $obj^2$ to original desired goal $dg^1$ and $dg^2$ respectively. Thus, the $g^{grip}$ is replaced with $\textbf{0}^3$, $g^{obj^1}$ is $dg^1$, $g^{obj^2}$ is $dg^2$, the $g^{obj^3}$ is also $ag^3$ to keep $obj^3$ in position.

Obviously, according to the above encoding rule, $e_g^{(j)}$ depends on the number of objects, i.e. $N$ and specific task index, i.e. $(j)$.  What's more, the encoding of two-object is a simpler version of three-object, and the single-object is also a simpler version of two-object. For brevity, we only give the typical formulation here, for $N>6$, $6<j<2N-4$, and $i=(j+1) \mid 2$, the $e_g^{(j)}$ can be formulated as (\ref{eq-eg}):

\begin{equation}\label{eq-eg}
	e_g^{(j)}=\left\{
	\begin{aligned}
		&[ag^i||{dg}^{1}...{dg}^{i-1}||\textbf{0}^3||ag^{i+1}...ag^N],   && \text{if~}j=2i-1,\\
		&[\textbf{0}^3||{dg}^{1}...{dg}^{i-1}||{dg}^i||ag^{i+1}...ag^N], && \text{if~}j=2i,\\
	\end{aligned}
	\right
	.
\end{equation}

Such a multi-goal \& multi-task encoding serves as the foundation of the following data transfer and self-guided exploration.

\subsection{Maximize the Use of All Data by HER}
% 这个章节主要是阐述，如何利用HER，最大化利用不同策略探索到的数据。
%当有了一个多目标多任务的网络时，探索过程中，不仅会有大量的数据，还有不同的策略的数据。
When a network that can handle these multi-goal \& multi-task is developed, it will generate plenty of data during the process of exploration, not only from failure experiences but also from different policies.

% 对于持续学习来说，主要就是保证高效率的知识传递。而对于我们RHER框架来说，不仅可以利用它自己探索的数据，通过修改目标，还能利用其他策略探索到的数据。
The CL methods aim to transfer knowledge effectively. In the RHER framework, updating a policy can not only use its own explored data but also relabel the data collected by other policies by HER. 

% 具体来说，如果为了更新策略pi^j，我们在各种策略混合探索的轨迹中，采样一个transition，然后将原始的目标egt，0.2的概率修改成ej_gt,0.8的概率修改成后视经验目标，参考公式
Coincidentally, for continual RL, the agent also needs to generate non-negative samples by HER. Specifically, to update the policy $\pi^{(j)}$ of $\psi_j$, a transition $(s_t, a_t, s_t', s_{dg}=e_{g_t})$ will be sampled from a trajectory jointly explored by different policies. Then the original goal encoding $e_{g_t}$ will be replaced by $e^{(j)}_{g_t}$ as (\ref{eq-eg}) with probability 0.2, and the hindsight goal as (\ref{eq-eg-next}) with probability 0.8:

%the probability 0.8 were replaced by hindsight goal as (\ref{eq-eg-next}):
\begin{equation}\label{eq-eg-next}
	e^{(j)}_{g^h_{t}}=\left\{
	\begin{aligned}
		&[gp^i_{t+k}||{ag}^{1}_{t+k}...{ag}^{i-1}_{t+k}||\textbf{0}^3||ag^{i+1}_{t+k}...ag^N_{t+k}], && \text{if~}j=2i-1,\\
		&[\textbf{0}^3||{ag}^{1}_{t+k}...{ag}^{i-1}_{t+k}||{ag}^{i}_{t+k}||ag^{i+1}_{t+k}...ag^N_{t+k}], && \text{if~}j=2i,\\
	\end{aligned}
	\right
	.
\end{equation}

where $gp^i_{t+k}$ is the $t+k$ time step gripper position, $t<k<T$, and the $T$ is the trajectory length, which follows the $future$ strategy from \cite{HER1}.

Therefore, the RHER scheme can realize both the forward and backward data transfer together for CL, and alleviate the negative sparse reward problem by HER.\label{Section IV.C}

\subsection{Self-Guided Exploration Strategy (SGES)}

% 相比普通的持续学习来说，持续强化学习不仅需要分享数据，还需要从环境中采集新的有价值数据，甚至需要在线探索来消除离线数据带来的累计推断误差。
Compared with standard CL, continual RL not only needs to share data but also needs to actively interact with the environment to obtain new valuable data, and even needs online exploration to eliminate the accumulation of errors of inference caused by offline data. 

% 更重要的是，如简介说的，由于NNSR问题，有价值的数据
What's more, as mentioned in Section (\ref{Section I}), due to the NNSR problem, the agent needs to collect more useful samples, which denotes that the corresponding achieved goals need to be changed by the agent. Therefore, an efficient and stable exploration strategy is required. 

Inspired by the idea of a relay, when a traveler needs to explore further, he/she needs to be escorted by some experts, then he/she can quickly pass through the area that the expert is familiar with, and finally explore new areas by himself/herself. Also like students for scientific research, who are guided by advisers and other researchers until they need to explore a new field.

% 对于机器人操作任务来说，如果直接学习一个操作任务，那么它的探索复杂度将是指数级的，因此我们将从最简单的任务，reach到第一个物块，开始学习.幸运的是利用HER可以快速学会这个任务，agent将会获得第一个guide-policy，这个引导策略将会带领agent到达第二个阶段，有了第二个阶段的数据，p2也将会很快学会。 甚至于如果学习策略将智能体带偏了，引导策略也能及时的纠正过来。
%%%%%%%%%%%%%%%%%%%%%%%%%%%%%%%%%%%%%%%%%%%%%%%

\begin{figure}[h]
	\hspace*{-2mm}
	\includegraphics[width=0.9\textwidth]{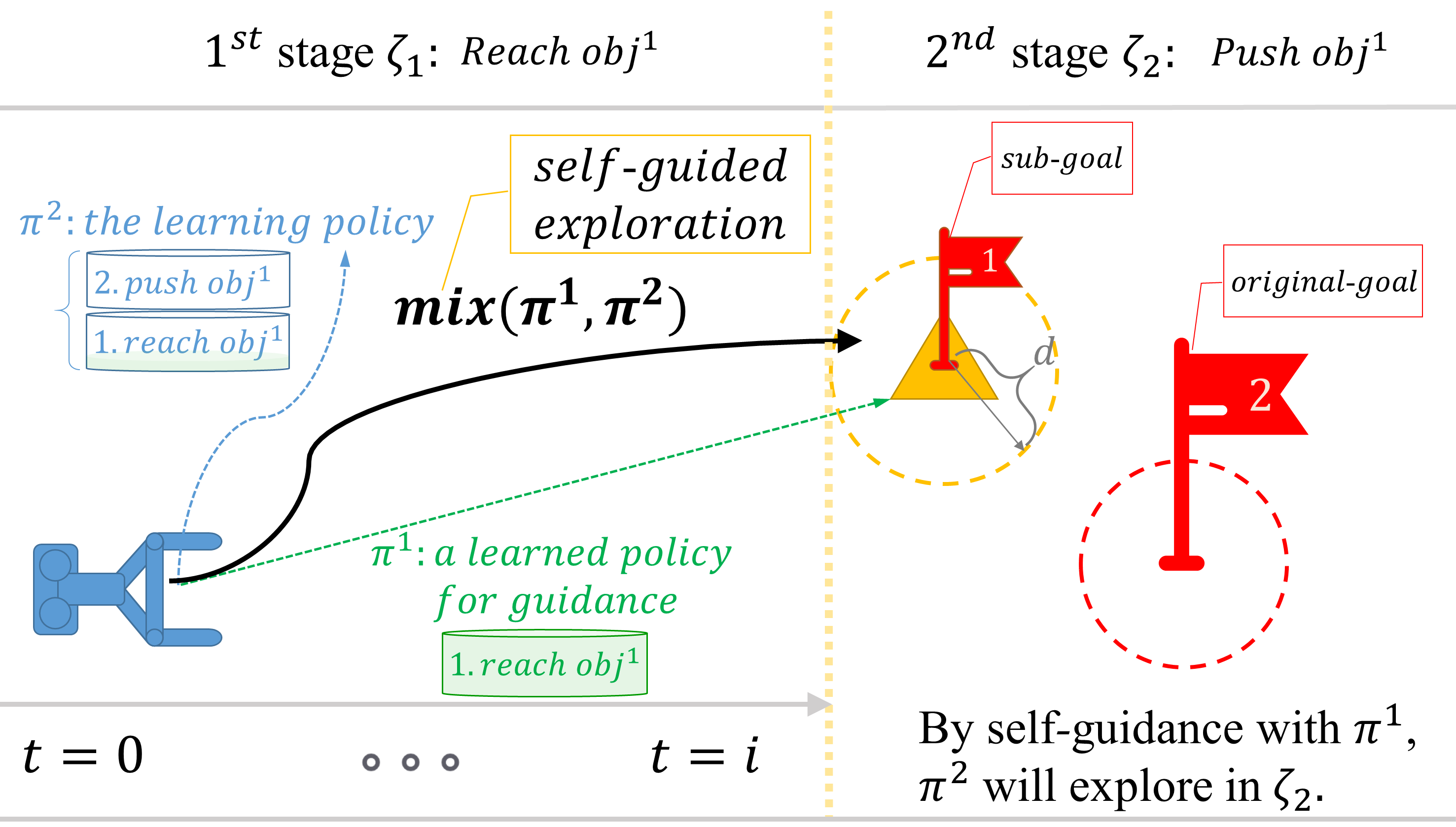}
	\caption{Illustration of Self-Guided Exploration Strategy (SGES) in a toy push task.}
	\label{Fig-RHER-SGES}
\end{figure}

As for robotic manipulation tasks, the RL agent can be guided by itself with RHER. The agent learns the simplest sub-task quickly by HER because the corresponding achieved goals can be changed by the agent all the time. We use this learned policy as a guide-policy, and the next more complex one as a learning-policy. RHER mixes the learning-policy and guide-policy in the sub-task that the latter has learned. Even though the learning-policy deviates a little from the optimal trajectory at first, the guide-policy can correct it in time. Thus, the agent can quickly reach the next stage to collect new valuable data for the learning-policy. 

An example diagram can be found in Fig. \ref{Fig-RHER-SGES}. The learned reach policy $\pi^1$ can directly reach the object in $\zeta_1$, and the trajectory of a learning policy $\pi^2$ is far away from the object. When $\pi^1$ and $\pi^2$ are mixed in $\zeta_1$, the trajectory is corrected to reach the object. Therefore, the agent can quickly collect valuable samples for $\pi^2$ by this self-guidance.

% 把related work的内容贴到这儿了，待会儿再详细修改。
The closest related work is the Residual-RL, which expects to combine the strengths of a model-based controller and Deep RL in two different ways. The first way is to superpose the signals of both policies \cite{Res-add3, ral-res-add4}. What these studies have in common is that the guide-policies are fixed and contain prior knowledge. The second way is to select a policy according to the certain probability \cite{Res-prob1, Res-prob2}, which is suitable for this work because the guide-policy will be updated online. In this manner, we mix policies with the same probability in SGES. To the best of our knowledge, we present the first method to realize self-guidance exploration without any pre-defined controller.

Such a self-guidance exploration strategy introduces the second hyper-parameter, i.e., the probability $\alpha$ of the guide-policy. A detailed ablation study will be conducted to show that this hyper-parameter is also insensitive (see Section \ref{Section VI.D}). Specifically, for exploration in RHER, the current stage index $j$ can be checked by (\ref{eq-stage}) and the action is selected according to (\ref{eq-SGES}):
\begin{equation}\label{eq-SGES}
	a=\left\{
	\begin{aligned}
		&{\pi}^{r} &&  \text{with/prob.}\ \beta,\\
		&{\pi}^{max(g, j)} && \text{with/prob.}~\alpha,\\
		&{\pi}^{min[max(g, j)+1, 2N]} && \text{with/prob.}\ 1-\alpha-\beta,
	\end{aligned}
	\right
	.
\end{equation}

where ${\pi^r}$ is a random policy, the probability of random actions is $\beta$, $\alpha$ is the guide probability, and $g$ is the guide stage index which means the corresponding test success rate of $\pi^{g}$ beyond a pre-defined threshold $sr$. In this work, for simple single-object tasks, $sr=1.0$, for complex multi-object tasks, $sr=0.8$. 

\begin{figure}[h]
	\hspace*{-2mm}
	\includegraphics[width=0.99\textwidth]{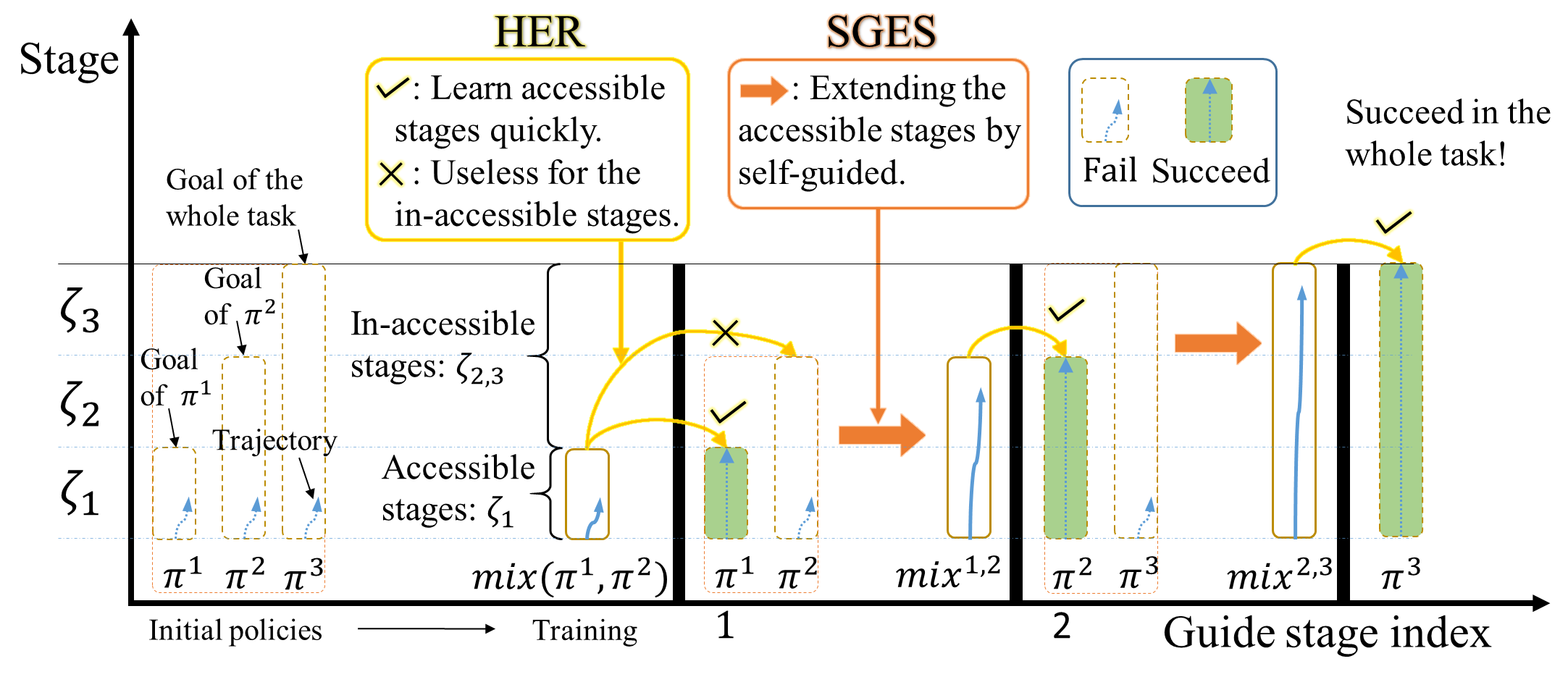}
	\caption{A diagram of relay policy learning for a task with 3 stages. By using HER and SGES, RHER can solve the whole sequential task stage by stage with sample efficient.}
	\label{Fig-RHER-relay}
\end{figure}

By the four components of RHER, the RL agent can solve the original complex sequential tasks with sparse rewards efficiently stage by stage. In short, by using HER, the agent can learn the accessible stages quickly, while by using SGES, the agent can extend the accessible stages quickly with self-guidance. A detailed relay policy learning diagram can be seen in Fig. \ref{Fig-RHER-relay}.

\label{Section IV.D}
%%%%%%%%%%%%%%%%%%%%%%%%%%%%%%%%%%%%%%%%%%%

\section{Experimental Setup}
In this section, the setup of six single-object, five multi-object simulations, and a real robot experiment are introduced to answer the following questions:
\newline
a) How does RHER compare with vanilla-HER in terms of efficiency and final performance? 
\newline
b) How sensitive is RHER to the two new hyper-parameter?
\newline
c) How well does RHER perform in preventing forgetting and transferring knowledge across tasks? 
\newline
d) Does RHER scale to more complex multi-objects tasks and a real-world scenario?

\subsection{Setup of the Simulation Environments}
\begin{figure}[h]
	\hspace*{-2mm}
	\includegraphics[width=0.99\textwidth]{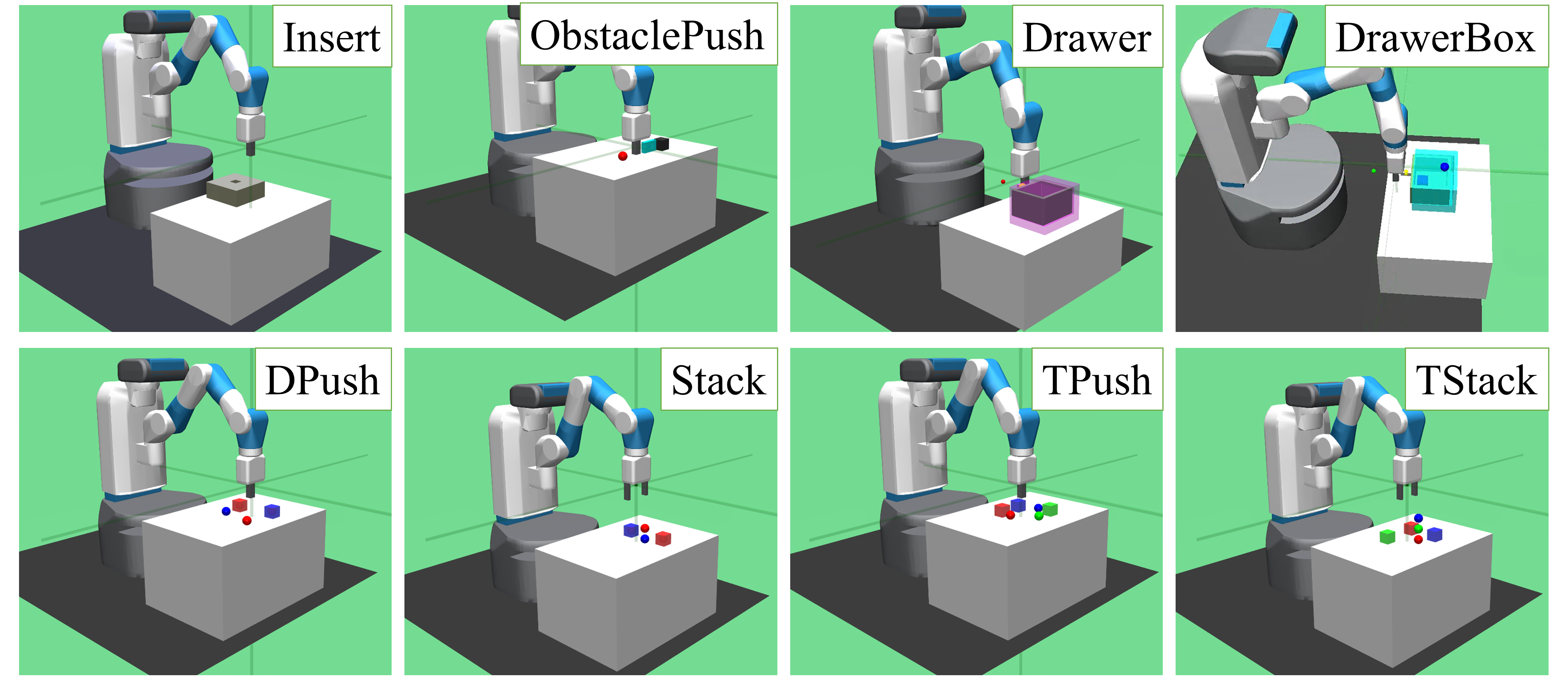}
	\caption{Different sequential object manipulation tasks. The single-object tasks are Insert, ObstaclePush and Drawer. The two-object tasks are DrawerBox, DPush and Stack. The three-object tasks are TPush and TStack.}
	\label{Fig-multi-obj}
\end{figure}

Three standard simulations are conducted in the gym fetch manipulation environments: FetchPush, FetchPickAndPlace, and FetchSlide. To illustrate that RHER can be generalized to other manipulation tasks, three additional single-object tasks are carried out. To illustrate that RHER can solve complex manipulation tasks, five multi-object tasks are conducted. These new tasks are shown in Fig. \ref{Fig-multi-obj}. The details are elaborated as follows. 

\noindent$\mathbf {Drawer:}$ 1) Reach the handler, and 2) slide the drawer to a target position by pulling the handler.\par

\noindent$\mathbf {Insert:}$ 1) Reach the hole, 2) insert a closed gripper into a hole with a random position. This paper creatively sets a virtual object in the inserting task, when the gripper reaches the vicinity of the hole, the virtual object will be moved by the gripper, so that HER can be applied.\par

\noindent$\mathbf {ObstaclePush:}$ 1) Reach the block, 2) push the block to a desired position in the presence of an obstacle that is unknown in size and shape. 

\noindent$\mathbf {DPush:}$ 1) Reach the first block, 2) push the first block to a desired position, 3) leave the first block and reach the second, and 4) push the second block to another desired goal.

\noindent$\mathbf {Stack:}$ 1) Reach the first block, 2) grasp the first block to a desired position, 3) leave the first block and reach the second, 4) grasp the second block on the top of the first block.

\noindent$\mathbf {DrawerBox:}$ 1) Reach the handler, 2) slide the drawer to a target position by pulling the handler, 3) reach the block, 4) grasp the block on the top of the drawer.\par

\noindent$\mathbf {TPush:}$ There are three blocks to push in order.\par

\noindent$\mathbf {TStack:}$ There are three blocks to stack in order.\par

The reward functions are still the most common binary rewards described in (\ref{eq-rew}). The other settings are referred to \cite{HER1} (details can be found in  \textbf{\href{https://github.com/kaixindelele/RHER/tree/main/gym}{\emph{Gym}}}). 

\subsection{Setup on a Physical Robot}
%Few studies directly use model-free RL to train the manipulation tasks with sparse rewards from scratch in realistic scenery. Even for the HER methods, the number of times to reset the environment will exceed $80$K \cite{HER1}, which is a tedious practical effort. However, RHER requires significantly fewer episodes of data to converge for a manipulation task such as the FetchPush. Therefore, this study builds a contact-rich push task, CuePush, to demonstrate that the model-free Deep RL has the potential to be trained directly in the real world.

\begin{figure}[h]
\hspace*{-2mm}
\includegraphics[width=0.9\textwidth]{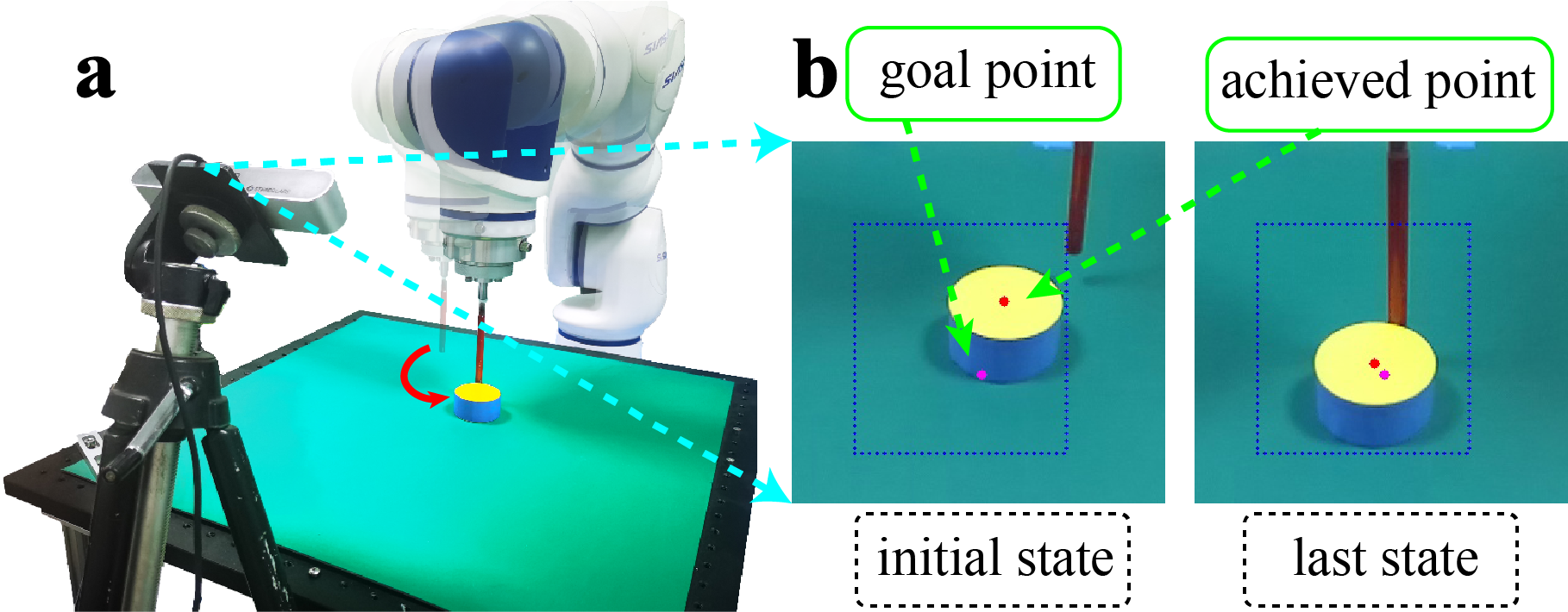}
\caption{CuePush: the real-world testing rig consists of a 7-DoF robot arm and a target cylinder. (a) A color-based camera detection system measures the Cartesian positions of the cylinder. (b) For the initial state, it samples a random point as the desired goal from the valid range.}
\label{Fig-real-env}
\end{figure}

\noindent$\mathbf {CuePush:}$ 1) reach a wooden cylinder, 2) push it to a desired position. 

As shown in Fig \ref{Fig-real-env}, the cylinder is 2.5 cm in radius and 3 cm in height, the end-effector is a metal cue that is attached at the end of a 7-DOF manipulator (SCR5, Siasun. Co, Shenyang, China), with the positions of the cue and the cylinder extracted by a monocular camera (ZED, Stereolabs, San Francisco, CA, US). This study even executes an open-loop automatic reset script to avoid human bias. Limited by the automatic reset script, the workspace of the robotic arm controlled by RL is only $9\times14$ cm. Therefore, the distance threshold $d$ in the simulation is reduced from 5 cm to 2 cm.

\subsection{Default Details of the Training Procedure}
\label{subsec-DetailsOfTraining}
% 其中batch size单多会有变化
For each episode, the agent performs 40 optimization steps on mini-batches of size 256 sampled uniformly from a replay buffer consisting of 1e6 transitions. The network structure and input scaling are default as \cite{HER2}. 

\textbf{Multi-processing training}: Since the sample efficiency and the performance of the HER increases with the number of parallel training processes \cite{HER2}, the proposed method is compared to the baselines in a multi-processing setting to validate its superiority. 

In this setting, 50 epochs are trained with 19 CPU cores (one epoch consists of $19\times 50 = 950$ full episodes), in which each episode consists of 50 steps.

\textbf{Single-processing training}:
%考虑到真实机器人的训练只有一个智能体，因此单进程训练的性能更值得关注。
Considering that there is usually only one robotic arm for online learning in the real world, we also make a comparison with vanilla-HER in a single-processing setting. Due to the lower sample diversity of single-processing, the vanilla-HER learns the tasks with 1200 epochs.

In this setting, the RHER and HER have 400 and 1200 epochs, a total of $400\times50$ and $1200\times50$ episodes, respectively, in which each episode consists of 50 steps. For the real environment, 15 epochs are set for training, and each episode just has 20 steps.

\section{Results}
% 这个章节要描述RHER相比于HER的性能和效果，以及利用三个消融实验，来证明每个组件的效果。用三个额外的单物块实验来证明RHER的通用性，用多物块实验，来进一步验证RHER可以对复杂的多物块操作任务也有良好的性能，最后在真机中测试，表明RHER具有不用仿真，直接真机从零学习的能力。
This section demonstrates the efficiency and effectiveness of RHER compared with baselines (HER). Ablation experiments will be conducted to illustrate the effect of each component or hyper-parameter. Three extra single-object tasks are carried out to show the generality of the RHER. Five more complex multi-object manipulation tasks further illustrate the broad application potential of RHER. Finally, the real-world experiment shows that the model-free RL can also be learned from scratch in real with sparse rewards. 

\subsection{Comparison with Baselines}
\begin{figure}[h]
\hspace*{-2mm}
\includegraphics[width=0.9\textwidth]{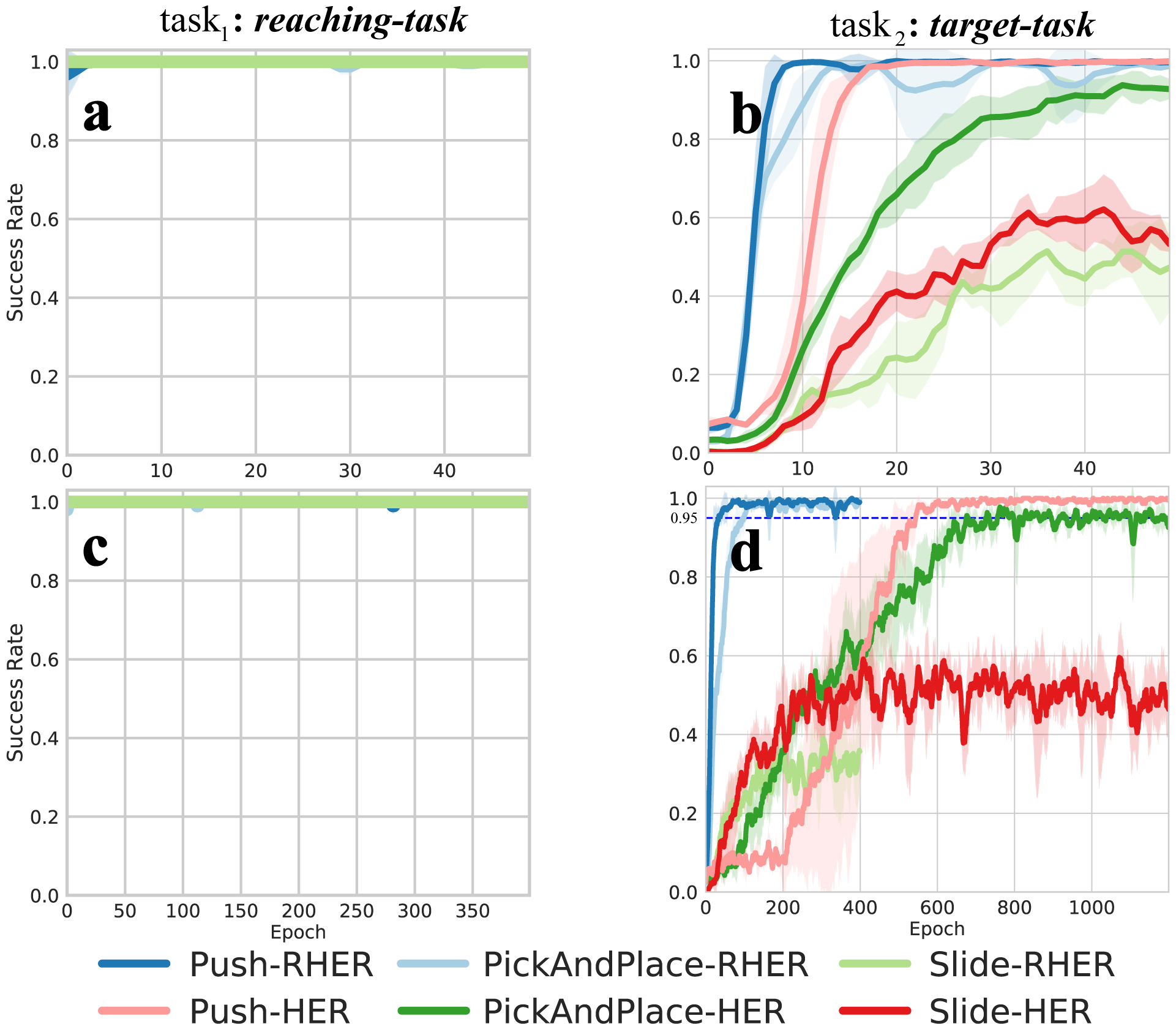}
\caption{Learning curves for the $\mathit{reaching{\mhyphen}tasks}$ and $\mathit{target{\mhyphen}tasks}$. Results are shown over 5 independent runs. The training curve represents the mean with standard deviation (based on 5 independent runs). (a) The $\mathit{reaching{\mhyphen}tasks}$ learning curves with multi-processing training. (b) Learning curves of $\mathit{target{\mhyphen}tasks}$ for multi-processing training. (c) Learning curves of $\mathit{reaching{\mhyphen}tasks}$ for single-processing training. (d) Learning curves of $\mathit{target{\mhyphen}tasks}$ for single-processing training.}
\label{Fig-baseline}
\end{figure}

% 第一个对比实验是，多进程和HER比较，取得了更高的采样效率，更好的性能，更低的方差。然而对于slide任务，效果不算太好。主要因为有两个，一个是slide任务对环境更加敏感，另外一个是这里的goal space超过了工作空间，不适合我们的方案。
\textbf{Multi-processing}: As shown in Fig. \ref{Fig-baseline}b, RHER achieves dramatically higher sampling efficiency, better performance, and lower variance (less sensitive to random seeds) in FetchPush and FetchPickAndPlace. However, in the FetchSlide task, the performance of RHER is slightly worse than that of vanilla-HER. This may be due to the FetchSlide task being more sensitive \cite{Yang-Joint}, and the goal space is beyond the workspace, which is not suitable for RHER.

% 单进程的实验结果更加凸显了我们算法的应用价值。具体来说，对于Push任务，RHER只需要31个epochs就能超过95%的成功率。比原始的HER效率高17倍！然而反直觉的是，单进程的采样效率比多进程更高。可能是因为后者需要在前期浪费了更多的探索机会。因此RHER具有更高的应用潜力
\textbf{Single-processing}: As shown in Fig. \ref{Fig-baseline}(d), in the single-processing cases, the HER exhibits quicker learning as well as more stable performance. To be specific, the method proposed needs only 78K interaction steps (31 epochs) to achieve a success rate of $95\%$ in the FetchPush task, demonstrating sample efficiency 17 times higher than that of the vanilla-HER (550 epochs). Counter-intuitively, it is shown that the single-processing cases have higher sample efficiency compared with the multi-processing because the latter wastes more sampling opportunities in the early stage. Therefore, the results show that RHER has more potential to be applied to realistic scenarios. 

% 对于持续学习来说，评价旧任务是否遗忘十分必要，如图ac所示，过去任务能很快的收敛且不会遗忘，且不需要其他的引导，不论是单进程还是多进程，这证明了我们多目标多任务编码可以让旧任务一直保持不忘。
\textbf{Continual RL}: For continual RL, it is necessary to evaluate whether the previous tasks are forgotten as the training process proceeds. As shown in Fig. \ref{Fig-baseline}(a), (c), the policies of the previous task ($\psi_1$) can converge quickly without guidance whether in the multi-processing or the single-processing case, which proves that the previous tasks are retained by the multi-goal \& multi-task encoding all the time.
\label{Section VI.A}

%第一个消融实验：在reach阶段如何进行策略组合。
\subsection{Ablation 1: Policy Combinations in $\zeta_1$}
\begin{figure}[h]
\hspace*{-2mm}
\includegraphics[width=0.95\textwidth]{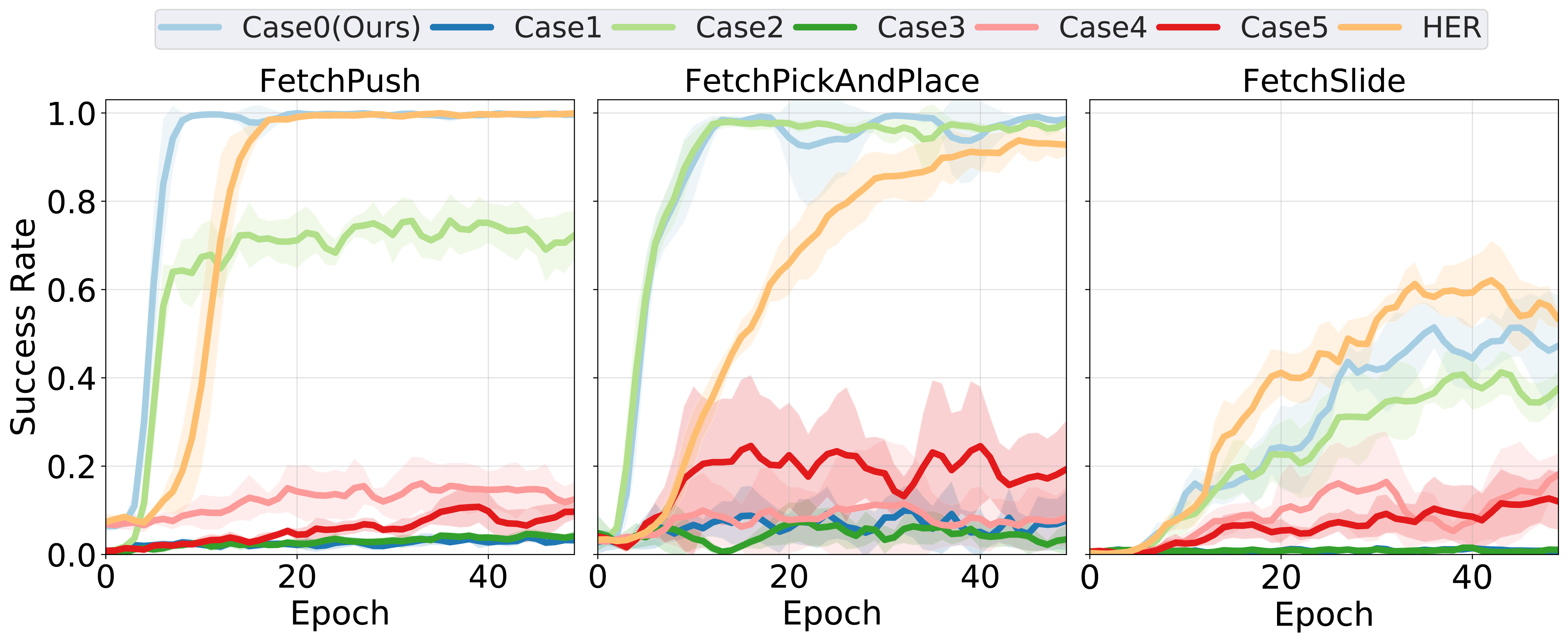}
\caption{Ablation study of task rearrangement. Details of these cases are described in Table\ref{table1}}
\label{Fig-policy-comb}
\end{figure}

%原始任务可以切分成2N个阶段，
\textbf{Comparison among Combinations}: First, according to the task decomposition (see Section \ref{Section IV.A}), the original task is divided into $2 (N=1)$ stages ({$\zeta_1$, $\zeta_2$}), but what kind of policy combination needs to be determined. As shown in Table \ref{table1}, for training or testing, $\zeta_1$ can be processed by $\pi^1$ and $\pi^2$ so that there are three policy combinations: $\pi^1$, $\pi^2$, and a mix of $\pi^1$ and $\pi^2$, denoted as $(\pi^1, \pi^2)$. When the $\zeta_1$ is processed only by $\pi^2$ in training, it is the same as the original HER, thereby excluding this option. There remain $2\times 3=6$ combinations and the corresponding results of these six cases are shown in Fig. \ref{Fig-policy-comb}. For example, for Case0, the agent mixes the $\pi^1$ and $\pi^2$ to explore the $\zeta_1$ in training, whereas $\pi^2$ in testing ($\zeta_2$ can only be solved by $\pi^2$). It can be seen from Fig. \ref{Fig-policy-comb}, that only Case0 (ours) takes into account both performance and efficiency. 

\begin{figure}[h]
\hspace*{-2mm}
\includegraphics[width=0.95\textwidth]{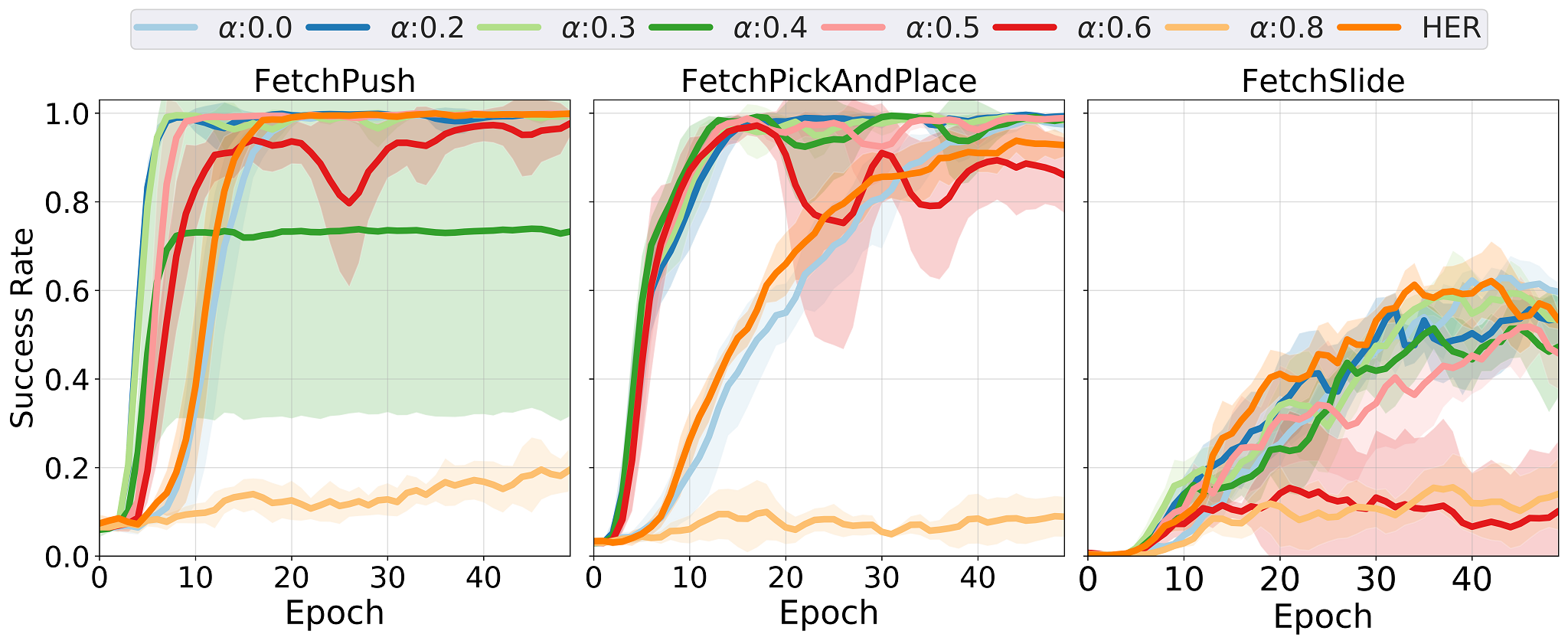}
\caption{Ablation study of exploration rate of SGES. Where the ratio of random policy $\pi^r$ is fixed at 0.2, $\alpha$ is the sampling probability of the guide-policy $\pi^1$ in $\zeta_1$.}
\label{Fig-SGES-rate}
\end{figure}

\textbf{Other Results}: There are some other interesting results in Fig. \ref{Fig-policy-comb}.  \textbf{1)} In Case4, $\pi^2$ does not explore $\zeta_1$ in the training process, so it cannot handle $\zeta_1$. This is largely attributed to the distributional shifts of actor-critic models using offline data. Details can be found in Fig. \ref{Fig-SGES-rate}, which shows that $\pi^2$ requires a certain percentage of exploration to correct bias from offline data. \textbf{2)} Combined with Case1, Case2, and Case0, the results show that, in testing, the performance of $\pi^2$ is damaged by other policies. The visualization results for the push task show that $\pi^1$ usually pushes the object out of the threshold, so the agent also takes $\pi^1$ for the next time step until the object fell off the table. The counter-intuitive result is caused because two policies cannot be toggled well. These results also explain the necessity of our task rearrangement formulation as (\ref{eq-task}).

Through the above experiments, an appropriate policy combination (Case0) that satisfies both efficiency and performance can be determined.\label{Section VI.B}

\begin{table}[h]\normalsize
 \caption{Different combinations of policies in the previous stage ($\zeta_1$).}
 \vspace{-3mm}
 \setlength{\tabcolsep}{0.3mm}	
 \renewcommand\arraystretch{0.8} 
 \begin{center}
  \begin{small}
   \begin{tabular}{c|c|c|c|c|c|c}
    \toprule
    Case & Case0(Ours) & Case1 & Case2 & Case3 & Case4 & Case5\\
    \midrule
  	 Training & $(\pi^1,\pi^2)$ & $(\pi^1,\pi^2)$ & $(\pi^1,\pi^2)$ & $\pi^1$ & $\pi^1$ & $\pi^1$\\
    Testing & $\pi^2$  & $\pi^1$ & $(\pi^1,\pi^2)$ & $\pi^1$ & $\pi^2$ & $(\pi^1,\pi^2)$\\
    \bottomrule
   \end{tabular}%
  \end{small}
 \end{center}
 \label{table1}%

\end{table}%

\subsection{Ablation 2: Distance Threshold}
\begin{figure}[h]
\hspace*{-2mm}
\includegraphics[width=0.95\textwidth]{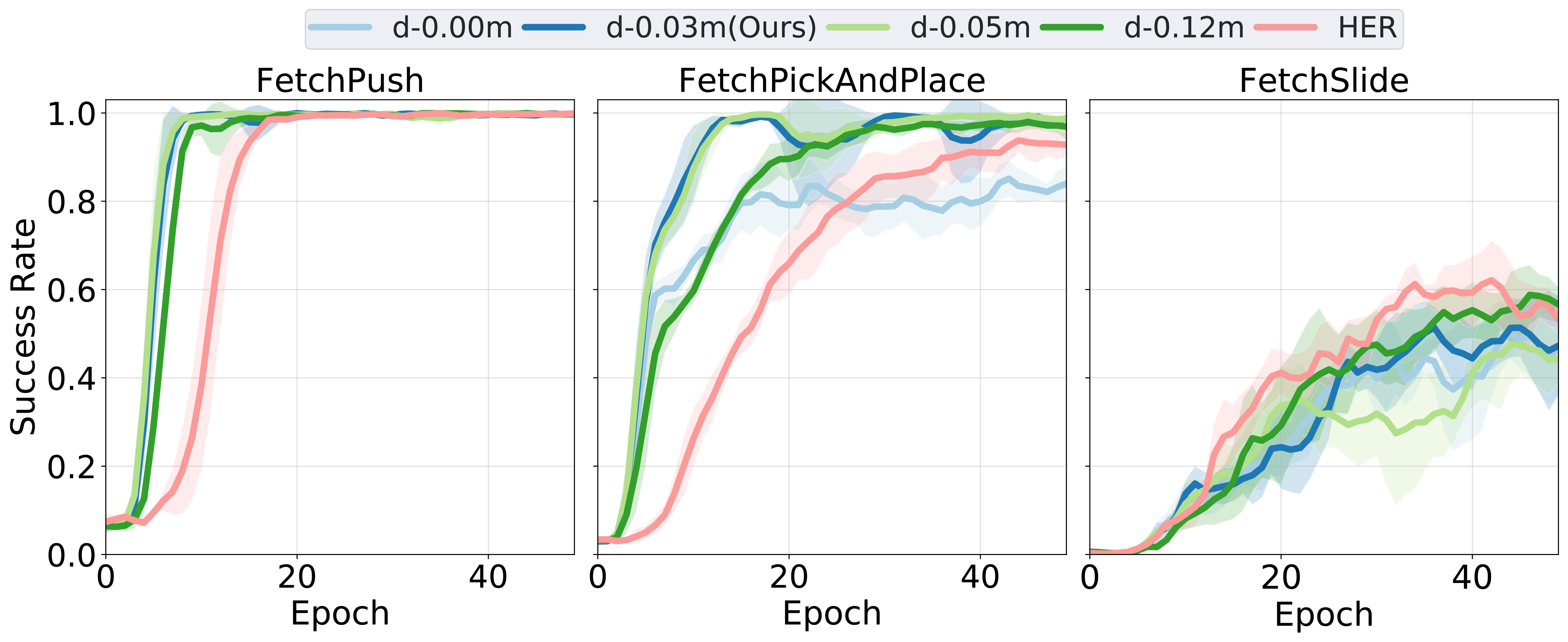}
\caption{Ablation study of different distance thresholds between gripper to the object.}
\label{Fig-dist}
\end{figure}

% 这个章节主要用于评估距离阈值对于性能的影响。
This subsection evaluates the effect of the distance threshold $d$ on performance. This is the first hyper-parameter introduced by RHER, which will affect the efficiency of exploration. However, Fig. \ref{Fig-dist} shows that if $d$ is too large ($d=0.12 m$), the performance will be degraded because the agent will also be hard to affect the achieved goals of the $\psi_2$. If it is too small ($d=0.0 m$), it will hinder from learning the $\psi_2$. However, as long as this distance is slightly larger ($d\in \{0.03 m, 0.05 m\}$) than the size of the object ($0.025 m$), the performance is highly sample-efficient. In other words, this hyper-parameter can be easy to choose. \label{Section VI.C}

\subsection{Ablation 3: The Rate of Guide-policy in SGES}
% 这个章节主要讨论阶段1，引导比例对结果的影响。
This subsection evaluates the guidance rate in $\zeta_1$, which is the second hyper-parameter in RHER. As shown in Fig. \ref{Fig-RHER-SGES}, in the reaching stage, there are three policies, the reaching policy $\pi^1$, the target policy $\pi^2$, and the random policy $\pi^r$. In this study, we fix the probability of $\pi^{r}$ as 0.2 and then increase the probability of the guidance policy $\pi^1$. From Fig. \ref{Fig-SGES-rate} it is clear that as long as there is guidance ($\alpha \in \{0.2, 0.3, 0.4, 0.5, 0.6\}$) by $\pi^1$, the target policy $\pi^2$ will be accelerated. It confirms that SGES is a crucial element that makes the exploration more effective. 

Interestingly, for two extreme cases, the experimental results can dispel some doubts. If there is no guidance from $\pi^1$ in $\zeta_1$ ($\alpha = 0.0$), the agent uses the offline data to train an auxiliary task, such as a $\psi_1$, which does not speed up the $\psi_2$. If there is no exploration of $\pi^2$ in $\zeta_1$ ($\alpha = 0.8$), the agent is unable to solve the $\psi_2$ by using offline data from $\pi^1$.\label{Section VI.D}

\begin{figure}[h]
\hspace*{-2mm}
\includegraphics[width=0.9\textwidth]{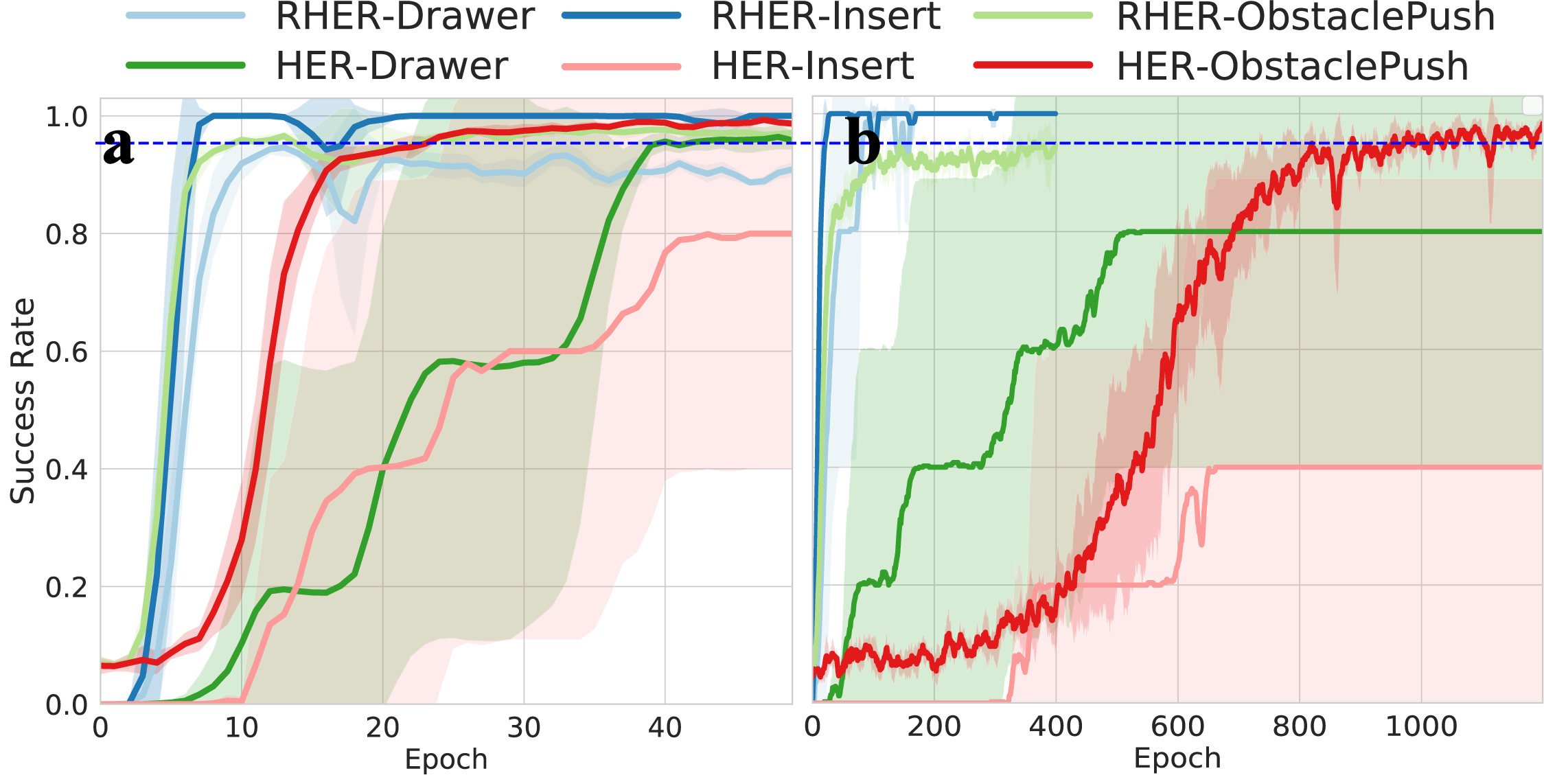}
\caption{Learning curves for three extended single-object tasks compared with vanilla-HER in (a) multi-processing and (b) single-processing.}
\label{Fig-extra-one}
\end{figure}

\subsection{Extending RHER to Three Extra single-object Tasks}
% 为了验证RHER的鲁棒性，我们将在三个额外的典型单物体任务中操作中验证。
To validate the robustness of RHER, three extended and typical single-object tasks are conducted with the same hyper-parameters as the above three classical tasks. As shown in Fig. \ref{Fig-extra-one}, in three challenging tasks, RHER still maintains high exploration efficiency and good performance, especially in single-processing. It is noted that RHER learns the Insert task from scratch within 65K time steps, which may have a positive impact on the industry. In contrast, vanilla-HER has a very high variance in all tasks, which means that some random seeds can be lucky to explore valuable data, while others remain stuck in local optima.\label{Section VI.E}

\subsection{Extending RHER to Five Multi-object Tasks}
% 为了验证自我引导探索和接力式学习的优越性，我们需要在更加复杂的多物体操作任务上测试RHER算法。
% 和一个物体的操作任务相比，多物体任务的复杂性显著增加，比如说两个物体的任务需要切分成四个阶段，三个物体需要切分成六个阶段。因此我们对之前的算法做了一个较大幅度的更新，在table2中列出了更新的细节。重要的更新有，深度学习库从tensorflow改成了pytorch, 强化学习算法从ddpg改成了td3, 为了让智能体快速学习最新的sub-task，我们将公式SGES中提到的st超参数从1.0改成了0.8，即选择已经学会的最复杂的sub-task policy作为引导策略，另外由于需要学习的样本多样性的提高，一个重要的超参数batch size需要从256提高到2048，最后由于任务难度的增加，我们相应的提升了episode length，即两物块为60，三物块的为70。
% 做了上述修改之后，我们在三个two-object 任务中做了测试。其中比较新颖的一个任务是drawerbox,它是需要先打开一个抽屉，然后从抽屉里面把物块拿到抽屉上面，这算是一种常见的challenging kitchen manipulation task。
% 另外还有更加复杂的的三物体操作任务，其中依次推三个物体到指定未知的任务，由于包含复杂的物体交互，很难设计一个简洁的传统控制器。
% 如图multi-object所示，我们RHER仍然可以快速的学会这些复杂的任务，而vanilla-her则连最简单的两物体任务都处理不了。另外可以注意到随着物体的增加，需要的回合数也会相应的增加。对于stack和push系列任务的上升曲线有明显不同，这可能它的任务特性有关。
% Therefore, by self-guided exploration and relay policy learning, RHER can address complex sequential object manipulation tasks with sparse rewards without any demostrations or model controllers.

To verify the effectiveness of self-guided exploration and relay policy learning, the RHER algorithm need to be test on more complex multi-object manipulation tasks.

Compared with the single-object tasks, the complexity of the multi-object task increases significantly. For example, the three-object tasks need to be divided into six stages, and it is also necessary to consider the relative position of other objects while moving one object. Therefore, we update the previous implement: 1) We used a PyTorch implementation of the TD3 \cite{TD3} method instead of the above the Tensorflow of DDPG. 2) To allow the agent to quickly learn the latest sub-task, we changed the $st$ mentioned in (\ref{eq-SGES}) from $1.0$ to $0.8$, which means that we select the most complex learned sub-task policy as the guide-policy. 3) Due to the samples with more diversity, the mini-batch size needs to be increased from 256 to 2048. 4) To stabilize the learned stages, we adapt the AAES methods from \cite{HRL}. 5) Finally, due to the difficulty of the multi-object tasks, we increased the episode length of two-object and three-object from 50 to 60 and 70 respectively.

After making the above modifications, we tested it on three two-object tasks as shown in Fig. \ref{Fig-multi-obj}. One novel task is the DrawerBox, which is a common challenging kitchen manipulation task. In addition, there are more complex three-object manipulation tasks. 

\begin{figure}[h]
	\hspace*{-2mm}
	\includegraphics[width=0.95\textwidth]{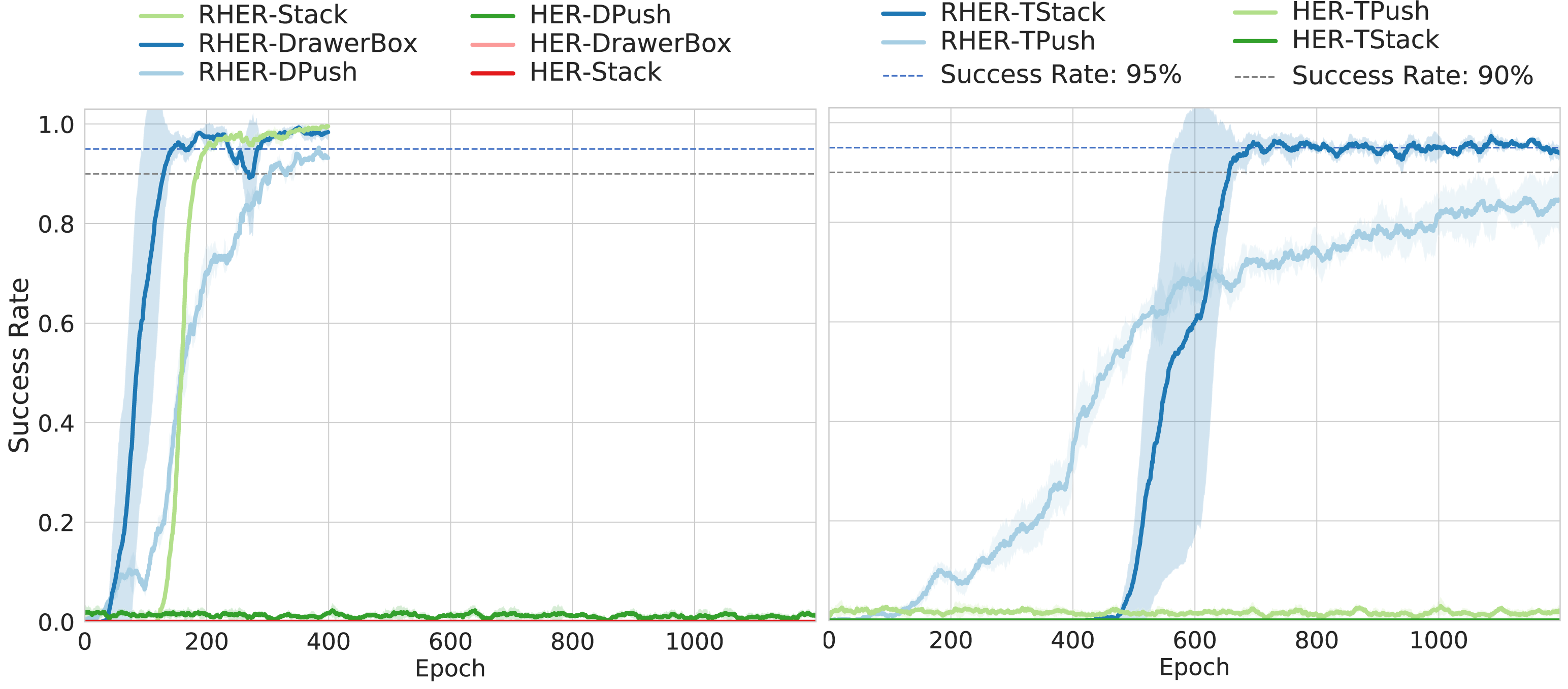}
	\caption{The curves of test success rate of multi-object tasks. The left and right parts show the results of two-object tasks and three-object tasks, respectively.}
	\label{Fig-multi_obj_exp}
\end{figure}

As shown in Fig. \ref{Fig-multi_obj_exp}, the RHER can still learn these complex sequential tasks quickly, while vanilla-HER can not learn even the simplest two-object tasks. Specifically, for two-object tasks, the success rate of RHER reaches $90\%$ less than $300 \times 50=15$K episodes, while the ACDER \cite{li2020acder} needs more than $150 \times 500=75$K episodes (as reported in their paper). As for TStack, RHER reaches an average success rate of $95\%$ after roughly $700 \times 50=50$K episodes. But for the TPush task, the final performance rises slower, because this task requires more complex trajectory compared to TStack.

In summary, by self-guided exploration and relay policy learning, RHER can address complex sequential object manipulation tasks efficiently with sparse rewards without any human demonstrations or pre-defined model controllers.
\label{Section VI.F}

\subsection{Learning on A Physical Robot from Scratch}

\begin{figure}[h]
	\hspace*{-2mm}
	\includegraphics[width=0.9\textwidth]{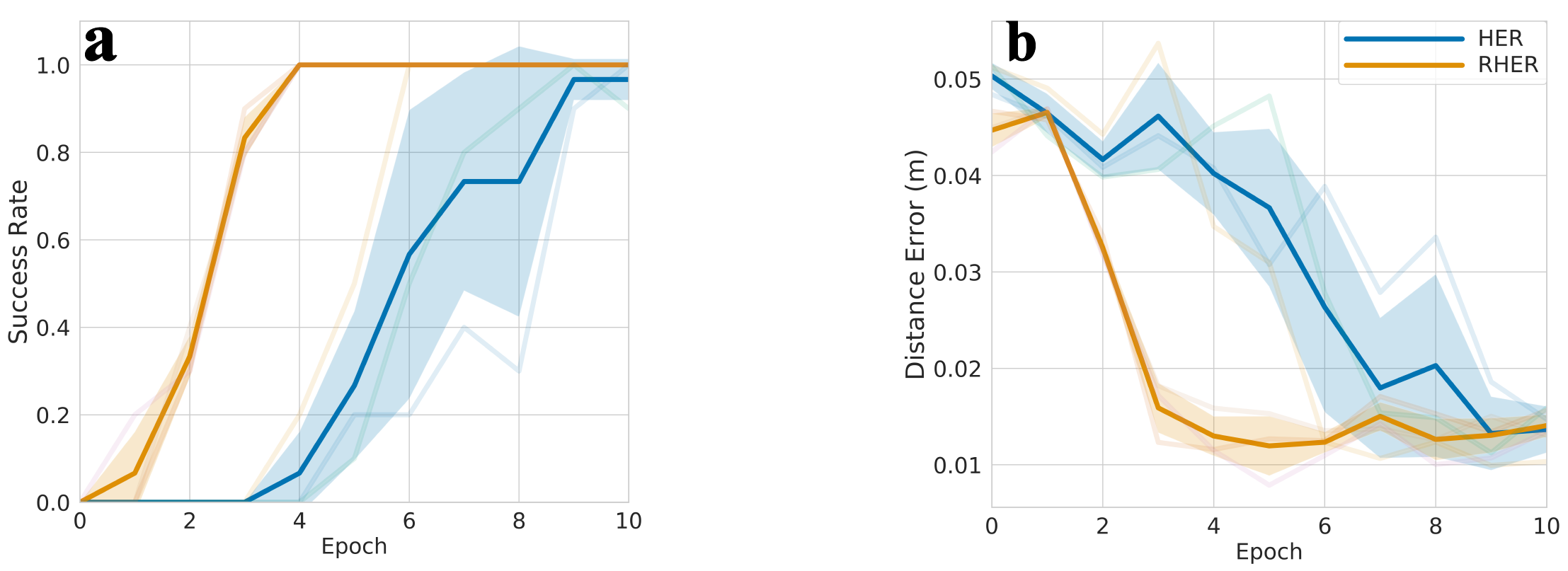}
	\caption{The curves of test success rate and distance error on CuePush.}
	\label{Fig-real}
\end{figure}

In CuePush, the end-effector is a smooth cue, and the object is a cylinder. As a result, the task is hard to design a suitable controller, and the automatic reset script usually pushes the object off the reset goal so that it needs to take multiple attempts. However, RHER learns this push task with the success rate of $10/10$ within 5K steps (250 episodes) and reduces the distance error to about 1cm, which is still efficient and stable, as shown in Fig. \ref{Fig-real}. If the workspace is larger, there will be more improvement compared with vanilla-HER.\label{Section VI.G}

\subsection{Comparison with State-of-the-art Methods}
To further evaluate the sample efficiency of RHER, a comparison of RHER with other HER-based state-of-the-art algorithms in the literature can be found in Table \ref{table2}. Since some papers do not release their code, this study uses the results of their paper for comparison. We compare the number of interactions required when achieving the same mean test success rate of $95\%$. Even though the IHER is a type of model-based method and ACDER has dynamic initial states to alleviate the non-negative sparse rewards, RHER still requires minimal interactions in two standard robotic tasks, as shown in Table \ref{table2}.

\begin{table}[h]\normalsize
 \caption{Number of interactions needed for test success rate $95\%$ in Push and Pick task with SOTA methods.}
 \vspace{-3mm}
 \setlength{\tabcolsep}{0.8mm}	
 \renewcommand\arraystretch{0.6} 
 \begin{center}
  \begin{small}
   \begin{tabular}{c|c|c}
    \toprule
    			& FetchPush($95\%$) & FetchPickAndPlace($95\%$) \\
    \midrule
	 vanilla-HER\cite{li2020acder} & 580K & 1680K\\
    \midrule
    ACDER\cite{li2020acder} & 110K & 260K\\
    \midrule
    IHER\cite{IHER} & $\sim$80K & $\sim$300K\\
    \midrule
	 RHER(ours) & \textbf{78K} & \textbf{250K}\\

    \bottomrule
   \end{tabular}%
  \end{small}
 \end{center}
 \label{table2}%

\end{table}%

\section{Discussion and Conclusion}
%本文首先讨论了一个有趣的问题，当设置一个过分远大的目标时（远到agent的行为无法改变结果），哪怕通过及时的反思，也不会对改进策略有什么帮助，必须要将任务拆解直到agent行为可以改变结果的地步，至于原始的复杂任务需要在子目标任务的手把手引导下，才能高效率的学会。, and it has to start simple and improve step by step。
%we first discuss a philosophical phenomenon: for an overly long-horizon task, the policy will not be improved effectively with hindsight when the agent can't change the target object. Therefore,

%我们提出的这种自我引导方式，可以很方便的迁移到其他任务中，比如一个稀疏奖励的复杂任务，但是可以设计一个非全局最优的dense奖励训练出来的策略，来引导稀疏奖励的策略快速学习。
In this paper, we propose a concise self-guided continual RL framework, called RHER, which solves complex sequential object tasks extremely efficiently with sparse rewards. The simulation experiments show that RHER can achieve stable and efficient performance on five single-object and five multi-object manipulation tasks, especially in the single-processing training. Finally, we trained an agent to complete the push task from scratch on a physical robot, requiring only 5K interaction steps. These results indicate that RHER is generalized to object manipulation tasks and that RHER has application potential for real tasks without building a corresponding simulation. It is worth mentioning that this relay-style idea can inspire multi-agent or hierarchical RL, and the self-guided idea can be extended to other tasks (e.g., using a policy trained with a sub-optimal dense reward function to guide a policy with a sparse reward function, etc.). In future work, we will extend our approach to other RL domains.

\bibliography{IEEEabrv,reference}
\end{document}